\documentclass[conference]{IEEEtran}
\usepackage{times}

\usepackage[numbers,]{natbib}
\usepackage{multicol}
\usepackage{graphicx}
\usepackage{subfigure}
\usepackage[bookmarks=true]{hyperref}

\usepackage{amsmath}
\usepackage{amssymb}
\usepackage{textcomp}
\usepackage{xcolor}
\usepackage{pgfplots}
\usepgfplotslibrary{groupplots}
\usepackage{tikz}
\usepackage{tikzscale}
\usepackage{cleveref}
\usepackage{subcaption}
\usepackage{float}
\usepackage{multirow}
\usepackage{makecell}
\usepackage{caption}
\usepackage{mathtools}
\usepackage{algorithm}
\usepackage[indLines=true,noEnd=true,rightComments=false]{algpseudocodex} 
\usepackage{siunitx}
\usepackage{comment}
\usepackage{booktabs}
\usepackage{alphalph}
\usepackage{wrapfig}
\usepackage{bbm}

\crefname{section}{Sec.}{Section.}
\Crefname{section}{Sec.}{Section.}
\crefname{figure}{Fig.}{Figure.}
\Crefname{figure}{Fig.}{Figure.}
\crefname{table}{Table.}{Table.}
\Crefname{table}{Table.}{Table.}
\crefname{equation}{}{}
\Crefname{equation}{Equation}{Equations}
\crefname{appendix}{Appendix}{Appendix.}
\Crefname{appendix}{Appendix}{Appendix.}

\newcommand{\ie}{\textit{i}.\textit{e}., }
\newcommand{\eg}{\textit{e}.\textit{g}., }

\DeclareMathOperator*{\argmin}{arg\,min}

\newcommand{\LEGO}{LEGO}
\newcommand{\apex}{APEX-MR}
\newcommand{\tool}{LT-V2}

\newcommand{\home}{HOME}

\begin{document}

\title{

\apex{}: Multi-Robot Asynchronous Planning and Execution for Cooperative Assembly

}


\author{
  Philip Huang$^{*, 1}$, Ruixuan Liu$^{*,1}$, Shobhit Aggarwal$^{1}$, Changliu Liu$^{1}$ and Jiaoyang Li$^{1}$\\
  $^*$Equal Contribution, $^1$Carnegie Mellon University \\
}


%

\maketitle

\begin{abstract}

Compared to a single-robot workstation, a multi-robot system offers several advantages: 1) it expands the system's workspace, 2) improves task efficiency, and, more importantly, 3) enables robots to achieve significantly more complex and dexterous tasks, such as cooperative assembly.
However, coordinating the tasks and motions of multiple robots is challenging due to issues, \eg system uncertainty, task efficiency, algorithm scalability, and safety concerns.
To address these challenges, this paper studies multi-robot coordination and proposes \apex{}, an asynchronous planning and execution framework designed to safely and efficiently coordinate multiple robots to achieve cooperative assembly, \eg \LEGO{} assembly.
In particular, \apex{} provides a systematic approach to post-process multi-robot tasks and motion plans to enable robust asynchronous execution under uncertainty.
Experimental results demonstrate that \apex{} can significantly speed up the execution time of many long-horizon \LEGO{} assembly tasks by 48\% compared to sequential planning and 36\% compared to synchronous planning on average.
To further demonstrate performance, we deploy \apex{} in a dual-arm system to perform physical \LEGO{} assembly.
To our knowledge, this is the \textit{first} robotic system capable of performing customized \LEGO{} assembly using commercial \LEGO{} bricks.
The experimental results demonstrate that the dual-arm system, with \apex{}, can safely coordinate robot motions, efficiently collaborate, and construct complex \LEGO{} structures. Our project website is available at \url{https://intelligent-control-lab.github.io/APEX-MR/}.

\end{abstract}

\IEEEpeerreviewmaketitle

\section{Introduction}

Multi-robot manipulation is critical in robotic applications, such as industrial assembly \cite{MICHALOS201081,BRUNELLO2025102920}, material handling \cite{SMITH20121340}, and object arrangement \cite{Gao2022-iu,Gao2024-bl}, etc. 
Compared to a single-robot setup, a multi-robot arm system can easily expand the system's overall reachable area.
Besides, with a team of robot arms, a task can be accomplished more efficiently \cite{Shome2021-ke} by having each robot execute individual tasks in parallel.
In addition to improving task efficiency, multi-robot systems can achieve significantly greater dexterity and are \textit{necessary} in many applications that require collaborations, \eg cooperative assembly, since certain tasks cannot be done with only one arm.

\begin{figure}
\centering
\includegraphics[width=\linewidth]{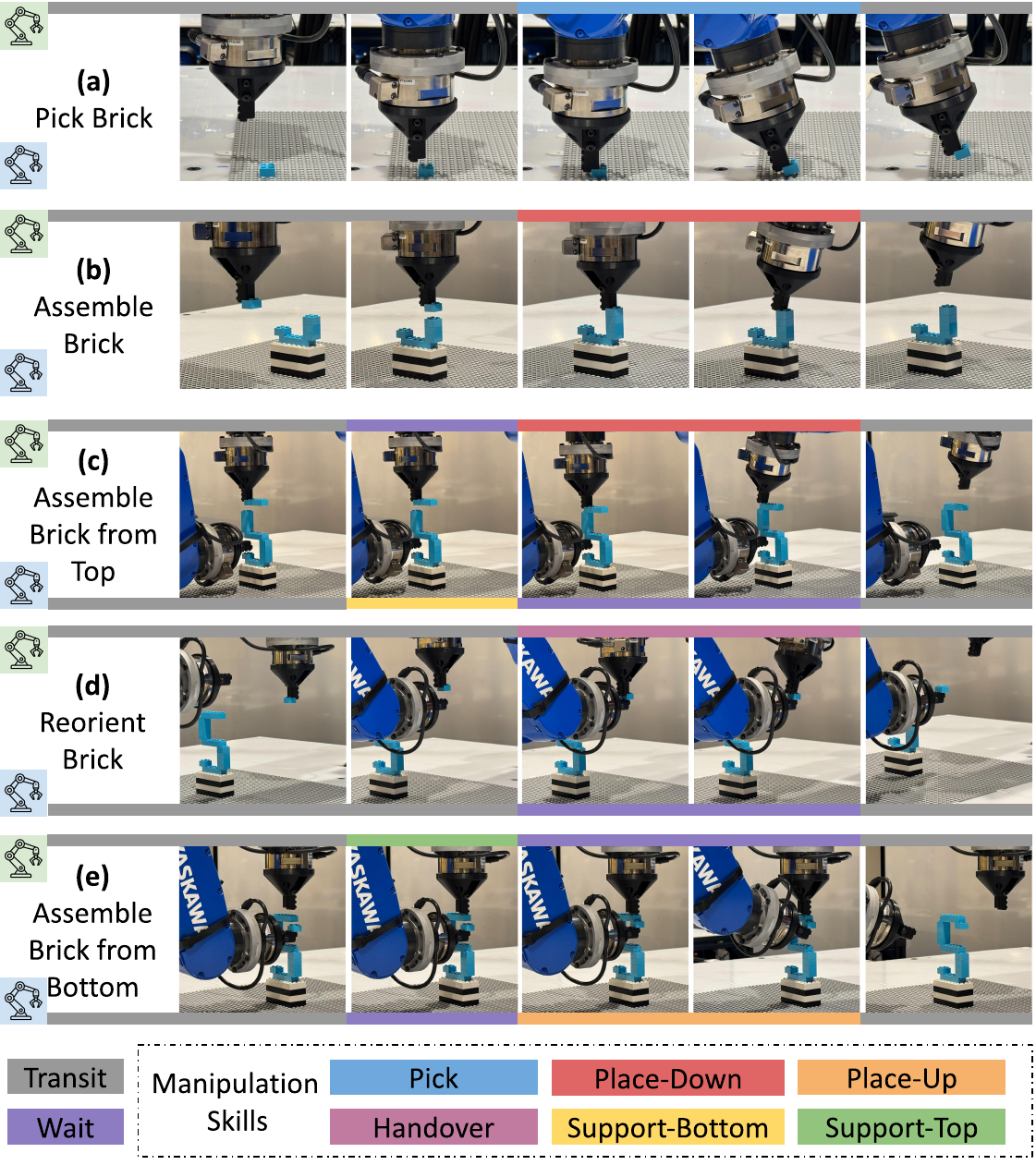}
    \caption{Illustrations of bimanual cooperative assembly. Manipulation skills denote contact-rich operations for assembly.
    \label{fig:dual_arm_example}}
    \vspace{-0.4cm}
\end{figure}

\LEGO{} assembly is an example of cooperative assembly.
In a \LEGO{} structure, the bricks are assembled by forcing the top knobs of a brick into the bottom cavities of another brick. 
The bricks are held together passively, \ie by the friction in the knob-to-cavity connections.
Thus, the connections are not rigid and subsequent manipulation, if performed inappropriately, can easily break the existing structure.
Due to the nature of the passive connection, a multi-robot system is necessary for such cooperative assembly.
\cref{fig:dual_arm_example} demonstrates example assembly operations in the construction of a \LEGO{} structure.
\cref{fig:dual_arm_example} (a) shows the robot picking up a brick by disassembling it from the \LEGO{} plate (\ie pick). 
\cref{fig:dual_arm_example} (b) illustrates the robot assembling a brick by placing it at the desired location and forcing a solid connection (\ie place-down).
One robot is sufficient for these two tasks since the manipulated object is fully supported and the operation would not collapse the existing structure.
On the other hand, \cref{fig:dual_arm_example} (c)-(e) showcase operations that require multi-arm collaboration.
In \cref{fig:dual_arm_example} (c), the upper robot is assembling a \LEGO{} brick on top of the character `S'.
To establish a solid connection, the robot needs to press down and force the knob insertion due to the passive connection nature.
However, since the existing connections are non-rigid, the place-down operation would break the existing overhanging structure.
Therefore, the lower robot is necessary to support and stabilize the structure from below (\ie support-bottom).
Similarly, in \cref{fig:dual_arm_example} (e), the lower robot assembles a brick from the bottom onto the character `S' by pushing it up and forcing the connection (\ie place-up).
The place-up operation would break the existing structure and therefore the upper robot is required to press down (\ie support-top) in order to stabilize the structure.
In addition to cooperative assembly, multi-robot collaboration also enables more dexterous object manipulation, \eg reorienting bricks in hand.
As shown in \cref{fig:dual_arm_example} (d), the upper robot initially grabs a brick from its top.
To have the robot grab the brick from its bottom, we can have the upper robot handover the brick to the lower robot.
With the ability to reorient objects in-hand, the robot can subsequently accomplish the assembly from the bottom as illustrated in \cref{fig:dual_arm_example} (e).
Despite being a toy brand, \LEGO{} has been widely used in entertainment, education, prototyping, etc.
It is ideal for assembly benchmarking because it is a low-cost, standardized, and highly customizable assembly platform.
Meanwhile, constructing \LEGO{} structures is a challenging contact-rich manipulation problem due to the high-precision requirement and non-rigid connections.
Thus, we use LEGO as our cooperative assembly benchmarking platform, and the remaining discussion is in the context of LEGO assembly.

Coordinating the tasks and motions of multiple robots to accomplish cooperative assembly is challenging for several reasons.
First, a system with more than a single robot introduces overhead and algorithmic challenges in their coordination.
As shown in \cref{fig:dual_arm_example}, certain operations involve contacts with objects (\ie pick, place-down, place-up, handover, support-bottom, and support-top) while some do not (\ie transit and wait).
There exist delays in controlling the robot and receiving sensor feedback, or even contingencies due to unexpected events when performing contact-rich operations, which would cause delays or stops to one or more robots.
The uncertainty makes it challenging to safely coordinate the robots throughout the assembly process.
Second, due to the need for collaboration, robots often need to operate closely to each other as illustrated in \cref{fig:dual_arm_example} (c)-(e).
Even when executing under uncertainty in a confined shared workspace, robots must always avoid collisions.
Third, it is desired that the task can be accomplished more efficiently by a multi-robot system.
Thus, robots must have a comprehensive understanding of the cooperative assembly task and plan optimized motions to reduce the completion time, instead of frequently stopping to avoid collisions.
Lastly, the multi-robot system should scale as the complexity of assembly design increases.
It is necessary for the algorithm to scale to larger and more complex tasks.

Many recent works have proposed methods for planning multiple robot arms for assembly \cite{Jiang2023-cg, chen2022cooperativeMRAMP}, rearrangement \cite{Gao2022-iu, Gao2023-kx}, or general manipulation tasks \cite{shaoul2024gencbs}. 
However, these planners often assume a sequential execution order and synchronously moving robots, where all robots start a task at the same time and wait for other robots to finish. 
We take a different lens and propose \apex{}, an \textit{asynchronous} planning and execution framework to coordinate multiple robots robustly, efficiently, and safely. 
Given an assembly task sequence, \apex{} leverages integer-linear programming (ILP) to distribute the tasks to robots and generate a sequential robot plan.
Most importantly, \apex{} extends the temporal plan graph (TPG) \cite{Honig2016-ts} to multi-robot-arm systems and post-processes the sequential robot plan for asynchronous execution.
Specifically, the TPG captures all dependencies between different robots' tasks and motions and generates a partial-order graph, which can significantly reduce unnecessary wait time and is robust to execution delay and uncertainty.
To highlight the applicability of our proposed algorithm within a full multi-robot assembly pipeline, we deploy the proposed \apex{} to a dual industrial arm system to construct complex customized \LEGO{} structures up to 258 objects in simulation and up to 47 objects in the real world. 
In summary, our contributions are as follows:
\begin{itemize}
    \item We extend the TPG execution framework to multiple robot arms and show that it enables robust and safe multi-robot asynchronous execution under uncertainty.
    \item We propose a sequential multi-robot task and motion framework that is complementary to the TPG execution.
    \item We demonstrate algorithm deployment and system integration for bimanual \LEGO{} assembly. To our knowledge, this is the \textit{first} robotic system that can accomplish flexible (\ie customized complex designs instead of simple pick and stack) assembly using commercial \LEGO{} bricks.
\end{itemize}
The rest of this paper is organized as follows: 
\cref{sec:relatedworks} discusses relevant works.
\cref{sec:prelim} deliberates the input that \apex{} consumes and the assumptions we have.
\cref{sec:method} introduces our proposed \apex{}, an asynchronous planning and execution framework for multi-robot cooperative assembly.
\cref{sec:results} shows the experimental results and demonstrates the deployment to a bimanual system for \LEGO{} assembly.
\cref{sec:discussion} discusses limitations and future works.
\cref{sec:conclusion} concludes the paper.


\section{Related Works}\label{sec:relatedworks}


\textbf{Multi-Agent Path Finding and Execution} Multi-agent path finding (MAPF) \citep{Stern2019-ck} studies how to coordinate a large team of mobile robots in a discretized world environment, often in 2D grid worlds representing warehouse settings \citep{Ma2017-iu} or predefined roadmaps \citep{Honig2018-zd}. 
State-of-the-art MAPF algorithms can plan near-optimal collision-free paths for hundreds of robots in seconds \cite{li2021eecbs}. 

However, these MAPF algorithms achieve such impressive efficiency at the cost of neglecting robot kinematics and execution uncertainty. As a result, there is growing research within the MAPF community focused on the efficient and robust execution of (imperfect) MAPF plans on real robots. 
One of the most widely adopted frameworks is the temporal plan graph (TPG) \citep{Honig2016-ts}, originally proposed to post-process MAPF plans to meet robot kinematics by enforcing passing orders at locations visited by multiple robots. TPG has since been extended to MAPF execution frameworks under uncertainty \citep{Ma2017-wl} and mobile robot coordination in warehouses~\citep{Honig2019-gr, Varambally2022-ux}. 
Recent advances introduce bidirectional TPG \citep{Su2023-jp} and switchable TPG \citep{Berndt2020-xq, JiangAAAI25}, which further enhance TPG by allowing flexible passing orders at certain locations. 

Given the success of TPG in coordinating mobile robots, we aim to explore its applicability in coordinating robot arms. A key distinction of traditional TPG versus our use case is that the robot kinematics is more complex and may change over time as the robot arm picks up different objects, which is a key focus in this paper.

\textbf{Multi-Robot Arm Motion Planning} Motivated by the rise of bimanual manipulation systems, many early works in the field study the problem of dual arm motion planning \citep{Smith2012-if}. 
One naive approach that scales single-robot motion planning methods to two or more robots is to plan in the composite joint space, with methods such as RRT-Connect \citep{Kuffner2000-hn}, BIT* \citep{Gammell2020-bd}, or graphs of convex sets \citep{Marcucci2021-yv}. 
However, due to the curse of dimensionality, these methods struggle to find high-quality solutions as the number of robots increases.
Other common strategies include building a composite roadmap from the Cartesian product of individual roadmaps \citep{Gharbi2009-rv}, coordinating the speed of individually planned motions \citep{Bien1992-ze}, or using prioritized planning \citep{Erdmann1987-hu}. 
More recently, more specialized multi-robot motion planners have been proposed that are based on roadmaps (dRRT \citep{Solovey2016-dz}, dRRT* \citep{Shome2020-hc}, and CBS-MP \citep{Solis2021-aj}) or using MAPF techniques \citep{shaoulmishani2024xcbs, shaoul2024gencbs}. 
Some approaches also use online planning or control techniques to generate motions in real-time.
\citet{Zhang_undated-hg} propose an online motion coordination technique, in which they plan all robots' paths offline and a pairwise collision matrix between robots. Then, the speed of each robot can be efficiently planned online to avoid collisions, even in the presence of execution delays.
However, their method cannot always find a feasible motion and relies on a task reallocation process to avoid deadlock.
Other techniques such as a distributed model predictive controller \citep{Gafur2022-yi} or a dynamical system approach \citep{Mirrazavi-Salehian2018-jm} can also be used to generate multi-robot arm motions in real-time, but are not tailored towards long-horizon planning tasks.

\textbf{Multi-Robot Task and Motion Planning}
Beyond motion planning, multi-robot arm task and motion planning (MR-TAMP) have been studied since the 1990s, \eg object pick and place \citep{Koga1994-ty}, and many of which are designed for a dual arm setting \citep{Harada2014-zp, Gao2022-iu, Gao2024-bl, Smith2012-if}. Similarly to our method, these methods take a two-stage approach in which they first generate a task plan with robot assignment and grasp poses, then search for corresponding motion plans with composite state space or prioritized planning. The authors of dRRT* have proposed extending dRRT* to multi-modal roadmaps with given possible pick-up and hand-off configurations and searching for a task and motion plan directly \citep{Shome2019-ak, Shome2021-ka}. Given that motion planning calls tend to be costly, another approach is to generate promising task plans first in a lazy manner and subsequently find corresponding motion plans and backtrack when required. This approach can be implemented with a greedy method \citep{Hartmann2021-kf}, with a mixed-integer linear program \citep{Shome2021-ke, chen2022cooperativeMRAMP,10552885} or through a satisfiability modulo theories solver \citep{Pan2021-ar}. However, a major drawback is that task planning often assumes motions to be synchronous, which can be suboptimal in practice. Our TPG framework is designed to be complementary to these synchronous tasks and motion planners, as it can relax the synchronicity assumption with post-processing and shorten the makespan of the overall plan. Another significant concern is that many MR-TAMP methods are designed for simple environments such as object pick and place, and their planned robot paths are executed in open loop or only in simulation. Our framework scales to more complex environments and is designed to integrate with feedback controllers and manipulation skills.

\textbf{Multi-Robot Arm Motion Execution} Most existing work executes their multi-robot task and motion plan on real robots in a synchronized way. Since the popular motion planning framework \textit{MoveIt} \cite{coleman2014reducing} does not natively support moving multiple robots asynchronously, some recent work has sought to address this and enable asynchronous execution. \citet{Meehan2022Asynchronous} adds a path reservation component and treats the entire path of a moving robot arm as a static obstacle when planning another arm's motion, which can be too conservative and cause deadlock. \citet{Stoop2023-im} uses a central scheduler to check if a new path collides with previously scheduled paths and executes it asynchronously if there is no collision. Otherwise, the new path waits for the conflicting previous path to finish before it can start. However, their methods do not account for execution delays, may wait longer than necessary, and directly modify the \textit{MoveIt} software stack. In contrast, our TPG formulation is robust to arbitrary delays by design, minimizes robot wait time, and can be implemented without modification to \textit{MoveIt}.


\textbf{Robotic LEGO Assembly} Automating LEGO assembly using robots is challenging due to the high-precision requirement, tiny sizes of LEGO bricks, and non-rigid connections in the structure.
Most of the existing works address the LEGO assembly problem in simulation \cite{popov2017dataefficient,9341428}, which cannot be generalized to physical assembly due to the lack of simulators to simulate the connections between LEGO bricks.
Recent works \cite{10.1109/ICRA.2019.8793659,8593852,liu2023simulation,Liu2023-ny} assume that the structure is fully supported and only consider placing bricks on top of others, which are limited to assembling simple structures.
\cite{7759340} considers assembly using customized brick toys, which does not apply to LEGO assembly. 
In this paper, we apply APEX-MR to a bimanual system to construct complex customized LEGO structures beyond simple stacking.

\begin{figure*}
    \centering
    \includegraphics[width=\linewidth]{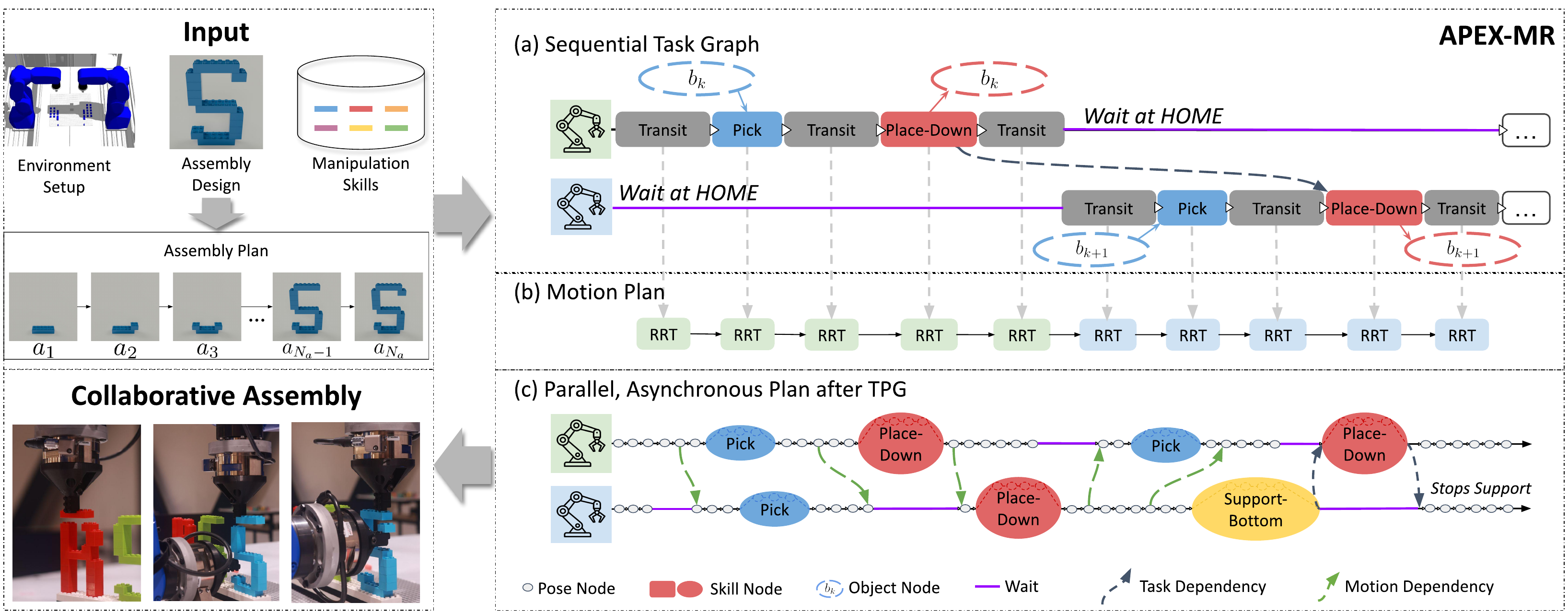}
    \caption{An overview of \apex{}. On a high level, \apex{} builds a sequential task plan given an assembly sequence, plans the motion of each task with RRT-Connect, and converts the solution to a parallel, asynchronous plan for execution with a TPG. Specifically, (a) shows the example of a task graph, (b) illustrates generating robot motion from the task plan, and (c) shows the example of a multi-modal TPG.}
    \label{fig:system_overview}
    \vspace{-0.4cm}
\end{figure*}

\section{Preliminaries}\label{sec:prelim}
To coordinate a multi-robot system to perform cooperative assembly tasks, we assume that three inputs are provided.

\textbf{Environment Setup}
We assume that the environment setup, including (a) geometries of all robots, (b) poses of objects to be manipulated $\mathcal{B}=[b_1, b_2, \dots, b_{N_b}]$, and (c) states of all obstacles, is known as shown in the input section in \cref{fig:system_overview}. 
We assume that the system consists of $N$ robots and $N_b$ is the number of objects that can be used for the assembly. 
An object $b_k$ is semi-static, meaning it can be grasped, attached, and moved by robots. 
Note that since duplicate objects are common in assemblies, $b_k$ and $b_{k'}$ can be identical objects, \eg identical $2\times 2$ bricks in the character `S' shown in \cref{fig:system_overview}.

\textbf{Assembly Plan}
Given an assembly design with $N_a$ objects, we assume that the assembly plan $A=[a_1, a_2, \dots, a_{N_a}]$ is provided as shown in the input section in \cref{fig:system_overview}.
Each step $a_j$ refers to an object, such as a $1 \times 2$ brick.
The assembly plan specifies the order in which each object should be assembled. 
Note that due to duplicate objects in $\mathcal{B}$, there are multiple candidates $b_i, \dots, b_k$ that can be used to accomplish an assembly step $a_j$ in $A$.
Furthermore, we require the assembly plan $A$ to be physically valid \cite{10.1145/3550454.3555525, Liu2024-hk}, ensuring that each task can be performed in reality. Specifically, the partially assembled structure after each step $a_j$ must be physically stable, and there exists at least one feasible grasp pose for each step. The assembly sequence also specifies whether each step $a_j$ requires two robots for cooperative assembly and the specific type of cooperation needed (\ie support or reorientation). Lastly, we assume that $\mathcal{B}$ is sufficient so that each object required for $a_j$ can be found in the environment.
More details on generating the assembly plan $A$ is included in \cref{sec:results} and \cref{sec:app-asp}.

\textbf{Manipulation Skills}
We assume that the robot manipulation skills are predefined and known.
We denote the skill set as $\mathcal{S}=[s_1, s_2, \dots, s_{N_s}]$, where $N_s$ is the number of skills an individual robot has.
For instance, a skill can be, inserting a pin, fastening a screw, picking up an object, etc.
In our case of \LEGO{} assembly, each skill is a composite of multiple motions parametrized by the pose of the manipulated object, the robot end-effector, and parameters learned from demonstrations. We assume that an algorithm, such as interpolation or RRT-Connect, can generate a reference robot path for the purpose of computing a collision-free coordination schedule. During execution, each skill is executed with a feedback controller, so that its exact execution time is unpredictable.
Examples of manipulation skills are shown in \cref{fig:dual_arm_example}, and more details are discussed in \cref{sec:results} and \cref{sec:app-skill}.

In addition to these inputs, we assume the existence of a collision-free \home{} pose for each robot that never blocks other moving robots from executing their tasks. Robots can also plan paths to move between different poses, or wait and hold their current pose in place, as shown in \cref{fig:dual_arm_example}.

\section{\apex{}: Asynchronous Planning and Execution for Multi-Robot System}\label{sec:method}

In this section, we introduce \apex{}, a framework for asynchronously coordinating the task plan, motion, and execution of a multi-robot system to accomplish cooperative assembly tasks. \cref{fig:system_overview} provides an overview of the three stages of \apex{}. 
The first task planning stage is discussed in \cref{sec:task_planning}.
Given the input discussed in \cref{sec:prelim}, a sequential task plan and the corresponding task graph, as shown in \cref{fig:system_overview} (a), are generated from the assembly using ILP. 
In the next stage, explained in \cref{sec:motion_planning}, the motion plan for each task is generated sequentially with a single-robot motion planner (see \cref{fig:system_overview} (b)). 
\cref{sec:asynch_exec} presents the last and most crucial stage, which converts the sequential plan to a multi-modal TPG (e.g. \cref{fig:system_overview} (c)) that can be executed safely, asynchronously, and efficiently in real robots.

For notation, we use $i\in[1, N]$ to index robots, $j\in[1,N_a]$ to index assembly steps in $A$, $m$ to index tasks for each robot, and $k\in[1, N_b]$ to index moveable objects in $\mathcal{B}$.

\subsection{Task Planning}
\label{sec:task_planning}
\begin{figure}[t]
    \centering
    \includegraphics[width=\linewidth]{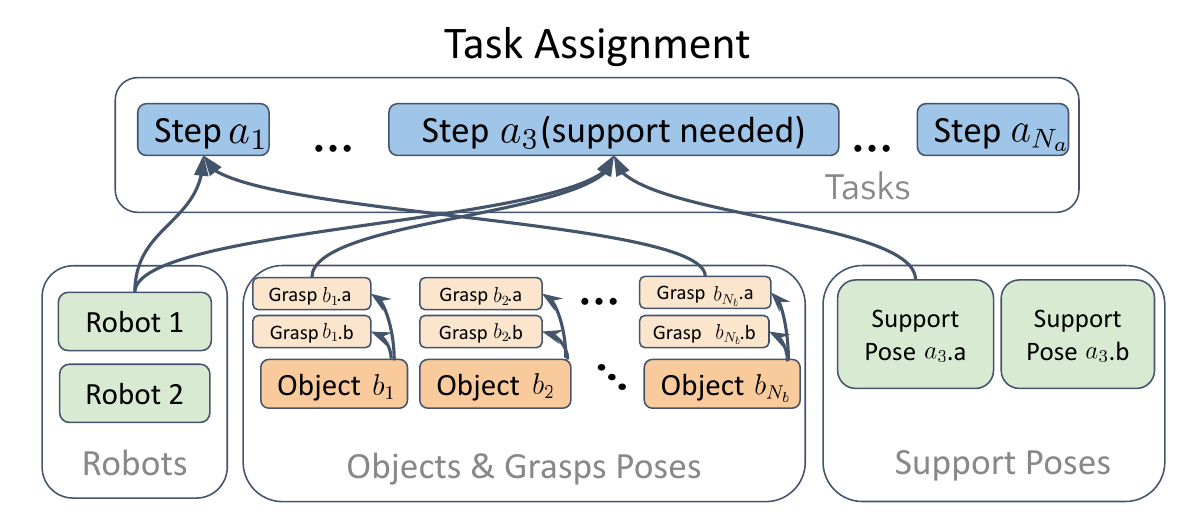}
    \caption{An overview of the task planning in \apex{}. Given the assembly sequence $A$, each step $a_j$ is assigned a robot, an object, a feasible object grasp pose, and a supporting pose if necessary. }
    \label{fig:task-planning}
    \vspace{-0.6cm}
\end{figure}

Given the assembly sequence $A$, task planning aims to construct a sequential task plan that includes robot assignment, robot target pose assignment, and object assignment. Specifically, for each step $a_j$ in the assembly sequence, the task planner must assign one responsible robot, a support robot if required, and an object $b_k$ of the correct type. In addition, the task planner must select the feasible grasp pose and a feasible support pose if required to perform the necessary manipulation skills for each $a_j$. \cref{fig:task-planning} provides an overview of all the decisions made. These decisions directly affect the feasibility of motion planning, as task planning determines whether there are collision-free and deadlock-free paths.

\textbf{Task Definition} We define a task $\mathcal{T}^i$ as a piece of work that requires a robot $i$ to either transit to a goal pose or to perform some manipulation skill $s$ to achieve a goal constraint. For example, a task may involve a robot arm picking up an object, supporting a structure, receiving an object handed from another robot, retracting to its \home{} pose, etc. 
In the context of \LEGO{} assembly, each assembly task $a_j$ is broken down into a sequence of tasks. As shown in \cref{fig:dual_arm_example} (a) and (b), a robot must first transit to the initial location of an object, pick the object, transit the object to the location for assembly, place down the object on the target structure, and finally transit back to the robot's \home{} pose. 
For an assembly step that requires collaborative assembly, the support robot is also assigned a sequence of tasks, such as transiting to the structure and supporting it in \cref{fig:dual_arm_example} (c) and (e). If a reorientation is required, as illustrated in \cref{fig:dual_arm_example} (d), the support robot is assigned a transit task and a pick task to collect the object, and a handover task to the other robot so that the object can be placed up to the structure. 

\textbf{Task Graph}
A task graph $\mathcal{G} = (\mathcal{V}, \mathcal{E})$ is a direct acyclic graph that represents the ordered set of tasks for all robots to complete the final assembly.
The task graph also implicitly represents the movement of manipulated objects, the evolving environment and planning scene, as well as the kinematics switches of the robots. A node is either a task node or an object node. A task node represents a task $\mathcal{T}^i_m \in \mathcal{V}$ for robot $i$, where $m$ is the index. Each task node also contains a robot target pose for this task. An object node represents an object $b_k$ and its pose. An edge from one task node to another in the graph, $\mathcal{T}^i_m \rightarrow\mathcal{T}^{i'}_{m'}$, represents a precedence constraint that states a dependency where $\mathcal{T}^i_m$ must finish before $\mathcal{T}^{i'}_{m'}$ can start. An edge from an object node to a task node, $b_k \rightarrow \mathcal{T}^i_m$, means that the object $b_k$ is kinematically attached to the robot $i$ at the beginning of this task. In contrast, an edge from a task node to an object node, $\mathcal{T}^i_m \rightarrow b_k $, represents that an object would be detached from robot $i$ after this task ends. 
Each robot $i$ has $M^i$ tasks. A task graph itself does not limit whether the tasks must be executed sequentially, synchronously, or asynchronously.

\textbf{Approach} The main idea of our approach is to find a sequential turn-based task plan according to the assembly sequence $A$, which is itself sequential. Only one robot is actively moving or executing skills at any time, while the other robot waits. Each robot returns to its \home{} pose at the end of completing an assembly step. 

Algorithmically, an ILP jointly optimizes robot assignment, object assignment, and target robot poses. A set of binary decision variables assigns a robot $i$ to an assembly step $a_j$ using the object $b_k$ and a feasible grasp pose. Another set of binary decision variables denotes the assignment of the support robot and support poses. We precompute feasible robot poses for grasping all available objects at initial and assembled positions, and support poses if necessary. The cost of assigning a robot $i$, an object $b_k$, and a corresponding grasp pose to an assembly step $a_j$ is estimated by the sum of the transit distances necessary for this step. The ILP finds the best set of assignments that minimizes the sum of costs to complete the assembly and an auxiliary term for load balancing while ensuring that the object type and support robot requirements are met. More details of the ILP formulation are discussed in \cref{sec:app-task}.

Combined with the precomputed robot poses, the optimized assignment gives a complete set of robot, object, and grasp pose assignments in each step. We then construct a corresponding sequential task plan and task graph (\eg \cref{fig:system_overview} (a)) with all inter-robot task dependencies and object relationships. A sequence of tasks for the assigned robot is added for each assembly step, and two object nodes are added to the task graph and connected to the pick task and place task nodes to indicate their attachment and placement, respectively. If a collaborative assembly is required, a sequence of support tasks is added to the support robot to be completed first. Inter-robot task dependencies are added to the task graph to constrain that the support task must precede any place task, and the following task after support can only start after the place task finishes. If two consecutive assembly steps $a_j$ and $a_{j+1}$ are assigned to different robots, a task dependency is added to ensure that the place task for $a_{j+1}$ must wait for the place task for $a_j$.

Compared to other MR-TAMP and assembly methods such as \citep{Gao2024-bl, Hartmann2021-kf, Pan2021-ar, chen2022cooperativeMRAMP}, \apex{} first generates a sequential plan. This has two advantages: (1) It is easy to reason about inter-robot collision for scheduling tasks and avoids expensive feasibility checks needed for parallel task execution, thereby making the ILP easier to solve; (2) The complexity of motion planning is significantly reduced since each task becomes a single-robot planning problem, which avoids solving a challenging and time-consuming multi-robot arm motion planning problem. 
It is worth noting that, while the sequential plan might seem inefficient for a multi-robot system, our TPG execution framework will post-process this plan to enable efficient parallel execution. 

\subsection{Motion Planning}
\label{sec:motion_planning}
Once the sequential task plan for each robot is determined, motion planning becomes straightforward. As illustrated in \cref{fig:system_overview} (b), \apex{} iterates over each task and uses a single-robot RRT-Connect algorithm to plan the path for this task when other robots are waiting. This is feasible because all other robots would be at a nonblocking stationary pose, \ie \home{}\footnote{We also assume that any robot waiting at a support pose is nonblocking.}. 
A reference path is also generated for tasks executed by specific manipulation skills based on the grasp/support pose.

The motion planner generates a planned path for every robot and every task. Each path $\tau^i_m$ for task $\mathcal{T}_m^i$ of robot $i$ is represented with a sequence of uniformly timestamped poses, $\{(C_n^i, t^i_n)\}, n \in [N^i_{start, m}, N^i_{end, m}]$ with step size $\Delta t$. $N^i_{start, m}$ and $N^i_{end, m}$ are the first and last index of the path $\tau^i_m$, $N^i_{start, 1} = 1$, and $N^i_{start, m+1} = N^i_{end, m} + 1$ for $m=1, \dots M-1$.
For each task, the planner first generates a sequence of poses ${C^i_n}$ for the robot $i$, then determines the corresponding timesteps.
To ensure that each robot takes turns to complete its task according to the task sequences, the initial timestep of each task is equal to the last timestep of the previous task in the sequential plan. Then, the timestep $t_n^i$ for each pose of the same task is set based on the maximum robot velocity, \ie $t^i_n = t^i_{n-1} + d(C^i_n, C^i_{n-1}) / v_{max}$. 
Since RRT-Connect may produce jerky and long paths, we use a randomized shortcutting algorithm to smooth suboptimal paths.
The output of motion planning should be a sequence of paths, one for each task, \ie $\tau^i_m,\; \forall i \in \{1, \dots, N\}, \; \forall m \in \{1, \dots, M^i\}$.

\subsection{Asynchronous Execution } \label{sec:asynch_exec}

From the sequential task and motion plan, \apex{} converts it to a TPG to improve the quality of the plan and support asynchronous execution. 
Importantly, the process to construct a TPG is not limited to a sequential plan and also applies to synchronous plans commonly seen in MR-TAMP works such as \citep{Pan2021-ar, Shome2021-ka, Shome2021-ke}.

\textbf{TPG Definition} \apex{} uses a multi-modal temporal plan graph (TPG) to represent an execution schedule for the team of robot arms. 
As illustrated in \cref{fig:system_overview} (c), multi-modal TPG is a directed acyclic graph $G=(V, E)$ with two types of nodes. A pose node $v_n^i$ corresponds to a configuration $C_n^i$ on the path of a transit task for robot $i$.
A skill node $\tilde{v}^i_n$ represents a manipulation skill (\eg pick, handover, support, etc.) that will be executed by some robot controller or policy. The skill node also contains a reference path generated in motion planning. 
A type-1 edge $(v_n^i \rightarrow v_{n+1}^i)$\footnote{Type-1 edge connects skill nodes too, \ie $(\tilde{v}_n^i \rightarrow v_{n+1}^i)$ or $(v_n^i \rightarrow \tilde{v}_{n+1}^i)$} connects two nodes from the same robot $i$ and indicates the order between two nodes. 
A type-2 edge $(v_n^i \rightarrow v_{n'}^{i'})$ represents an inter-robot precedence order that constrains robot $i'$ to wait for robot $i$ to reach $v_n^i$ (\ie reach pose $C_n^i$ or complete the manipulation skill) before moving to $v_{n'}^{i'}$. 
A type-2 edge can be added for both task dependencies and motion dependencies. 
For every pair of TPG nodes that could lead to collisions, this is a motion dependency, and a type-2 edge is required.
TPG ensures a deadlock-free execution schedule if and only if there are no cycles in the TPG.
In contrast to the TPG defined in \cite{Honig2016-ts}, a multi-modal TPG combines a TPG with a task graph and assigns a corresponding task $\mathcal{T}_m^i$ to each node $v_n^i$. Since there are kinematics switches and changes to the planning scene, each node also contains the robot kinematics (\ie attached object), which will be important for the TPG construction process.
 

\begin{algorithm}[t]
  \caption{Multi-Modal TPG Construction}
  \label{algo:mmtpg}
  \begin{algorithmic}[1]
    \LComment{Input: A task graph $\mathcal{G}$ and all robot paths $\tau^i_m$}
    \For{robot $i = 1, \dots, N$}
      \For {task $\mathcal{T}_m^i = \mathcal{T}^i_1, \dots, \mathcal{T}_{M^i}^i$}
          \If {task $\mathcal{T}_m^i$ is a transit to a goal pose}
          \State Add a sequence of nodes $\{v^i_n\}$ from $\tau_m^i$ 
          \State Add type-1 edges for consecutive nodes 
          \Else { Add a skill node $\tilde{v}^i_n$}
          \EndIf
          \State Add type-1 edge from $v^i_{{end, m-1}}$ to $v^i_{{start, m}}$
      \EndFor
    \EndFor
    \For {task dependency $(\mathcal{T}_m^i \rightarrow \mathcal{T}_{m'}^{i'}) \in \mathcal{E}$ such that $i \neq i'$}
        \State Add type-2 edge from $v^i_{{start, m+1}}$ to $v^{i'}_{{start, m'}}$
    \EndFor
    \For {robots $(i, i') = \{1, \cdots, N\} \otimes \{1, \cdots, N\}$}     
        \If{$i==i'$} continue
        \EndIf
            \LComment {(Optional) parallelize the following}
            \For {$n = 1, \dots, N_{end,M^i}^i$}
                \State Update robot $i$ kinematics if needed
                \For {$n' = 1, \dots, N_{end,M^{i'}}^{i'}$}
                \State Update robot $i'$ kinematics if needed

                \If {$t_{n'}^{i'} \geq t_{n}^{i}$ or $v_{n'}^{i'}$ precedes $v_n^i$ in $G$} 
                \State continue
                \EndIf
                \If {$v_{n'}^{i'}$ collides with $v_n^i$}
                    \State Add type-2 edge from $v_{n'+1}^{i'}$ to $v_n^i$
                \EndIf
                \EndFor
            \EndFor
        \EndFor
    \State Simplify TPG with transitive reduction

  \end{algorithmic}
\end{algorithm}

\textbf{Building a multi-modal TPG}
The process of constructing a TPG from a multi-robot motion plan can be interpreted as converting from the sequential robot paths to temporally dependent robot schedules. On a high level, inter-robot task dependencies are copied as type-2 edges for TPG, and type-2 edges for motion dependencies are constructed by scanning for collisions between all pairs of TPG nodes. A key benefit of this partial-order representation is that TPG does not specify a fixed time between two consecutive nodes, and thus allows execution delays. We outline the detailed procedure below.

The construction process begins with creating the nodes and type-1 edges. For each transit task $\mathcal{T}_m^i$, a pose node $v_n^i$ is constructed for each configuration $C_n^i$ of the path $\tau_m^i$.  
A skill node $\tilde{v}_n^i$ is created for every manipulation task.
The timestamp in the input path $t_n^i$ for each node $v_n^i$ is also recorded.
Each node $v_n^i$ is then connected to its successor $v_{n+1}^i$ by a type-1 edge, and the end node of a task $v_{end, m}^i$ is connected to the start node of the next task $v_{start, m+1}^i$.  When a robot waits at a node $v_m^i$, that is $C_m^i = C_{m-1}^i$, the node $v_m^i$ is removed because removing the wait action does not change the robot path or any temporal dependency in the TPG.

Next, all the type-2 edges are identified in the TPG. First, all inter-robot task dependencies $(\mathcal{T}_m^i \rightarrow \mathcal{T}_{m'}^{i'}) \in \mathcal{E}$ from the task graph $\mathcal{G}$ are added as type-2 edges to the TPG. 
Specifically, for every edge in the task graph $\mathcal{T}_m^i \rightarrow \mathcal{T}_{m'}^{i'}$, a type-2 edge is added from the beginning of task $\mathcal{T}_{m+1}^{i}$, $v_{{start, m+1}}^{i}$, to the beginning of task $\mathcal{T}_{m'}^{i'}$, $v_{{start, {m'}}}^{i'}$.
This type-2 edge ensures that robot $i$ only starts $\mathcal{T}_{m'}^{i'}$ after robot $i'$ finishes task $\mathcal{T}_{m}^i$.

Then, motion-level dependencies are identified by iterating over all pairs of nodes in the TPG with a double loop iterating over $(i, i')$.
Robot kinematics can be incrementally changed if node $v_n^i$ is the start of a new task that attaches or detaches an object.
For each node $v_n^i$, it is checked against all other nodes $v_{n'}^{i'}$ from a different robot ($i \neq i'$) and has an earlier timestamp $t^{i'}_{n'} \leq t_{n}^i$.
Collision checking for pose nodes means that the robot links and attached objects at the corresponding poses $C_n^i$ and $C_{n'}^{i'}$ are checked. 
For skill nodes, all robot poses on the reference path have to be collision-free simultaneously. 
If there is any collision between $v_{n'}^{i'}$ and $v_n^i$, a type-2 edge is added from the earlier node's successor, $v_{n'+1}^{i'}$, to the later node, $v_n^i$.
Setting the direction of type-2 edge based on timestamps in the sequential motion plan ensures that the multi-modal TPG has no cycles and thus is deadlock-free.
This way, if $v_{n'}^{i'}$ has a smaller timestamp than $v_n^i$ in the input sequential plan, \ie $t_{n'}^{i'} < t_n^i$, $v_{n'}^{i'}$ must still be executed before $v_n^i$ can start, avoiding any potential collisions.
When iterating $n'$, it is unnecessary to check collisions for any node $v_{n'}^{i'}$ that has a larger timestamp than $v_n^i$, \ie $t_{n'}^{i'} \geq t_{n}^{i}$, because those would be checked when the iterated robot pair ($i$, $i'$) are swapped.
Also, if the current $v_{n'}^{i'}$ is a predecessor of $v_n^{i}$ in the graph, $v_n^{i}$ already waits for $v_{n'}^{i'}$ and avoid collisions, so it becomes unnecessary to check them again.  

The total number of collision checks needed depends on the number of robots, discretized steps, and type-2 edges added from the task graph. Once every node has been checked against potentially colliding nodes, a transition reduction algorithm is applied, similar to \citep{Ma2017-wl} to simplify the TPG. A type-2 edge $v_n^i \rightarrow v_{n'}^{i'}$ is redundant if node $v_n^i$  is still a predecessor of node $v_{n'}^{i'}$ after the edge is removed. We remove all such redundant edges to reduce the total number of scheduling constraints and communication overheads during execution.

The primary bottleneck of the TPG construction process is the number of collision checks, which scales quadratically with respect to the number of robots and the number of nodes. 
To alleviate this, collision checking can be parallelized across many CPU threads, reducing its runtime. A pseudocode of the entire construction process is provided in Algo. \ref{algo:mmtpg}.

\begin{figure}[t]
    \centering
    \includegraphics[width=\linewidth]{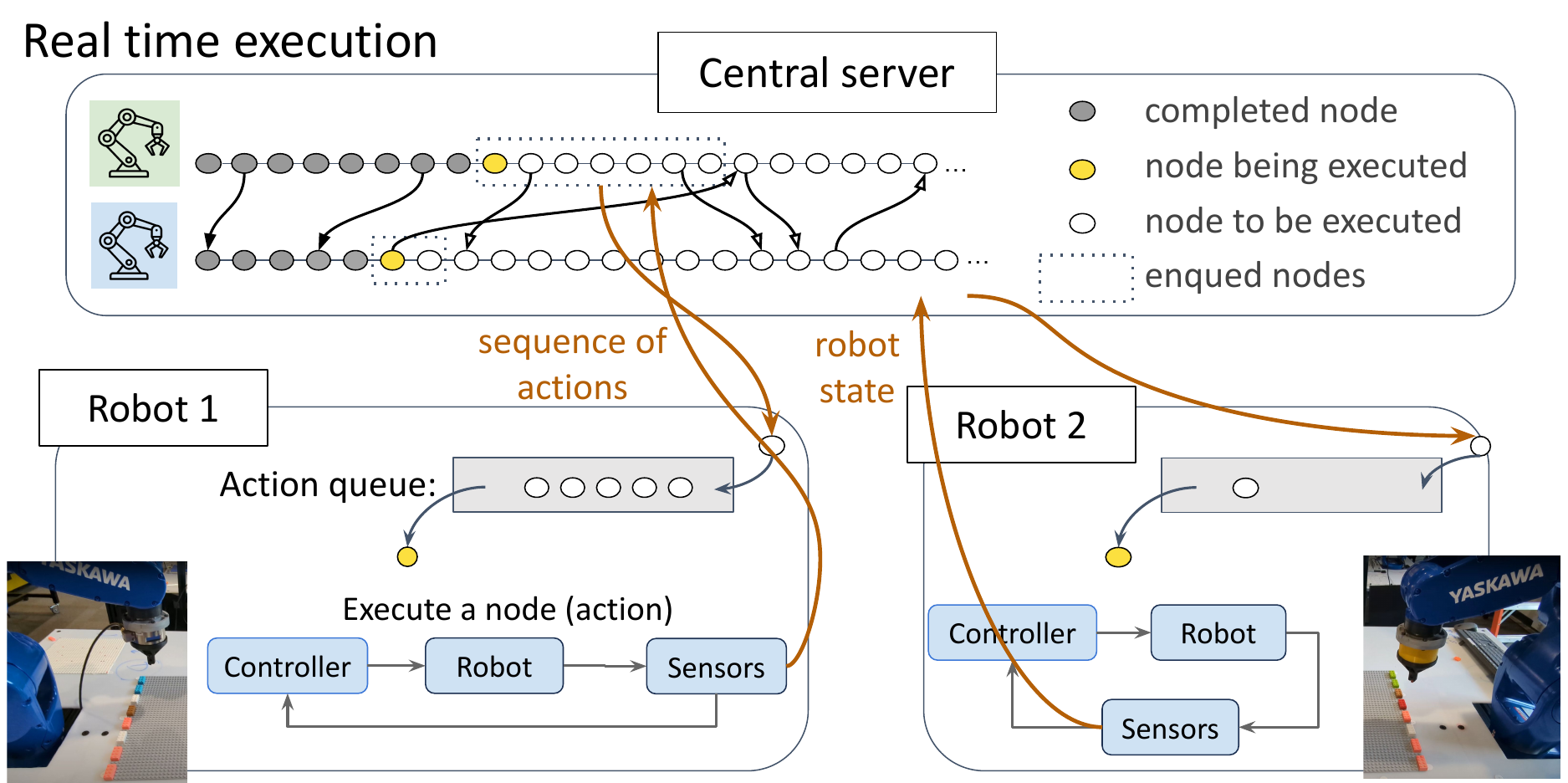}
    \caption{Illustration of the execution setup. TPG maintains and controls the execution schedule of all robots on a central server, gradually sending new path segments that can be safely executed. Each robot maintains a controller-sensing loop independently while updating its state with the central server.}
    \label{fig:tpg execute}
\end{figure}

\textbf{Further Optimization}
An optional step to further reduce the execution makespan and smooth the path is to skip the intermediate transition to \home{} after every assembly step. We use the following shortcutting algorithm, similar to the strategy implemented in \cite{Okumura2022-mw}, to achieve that while maintaining a collision- and deadlock-free plan. 

The anytime algorithm works by randomly sampling two nodes of the TPG ($v_n^i, v_{n'}^i$) from the same transit task $\mathcal{T}_m^i$, checking whether connecting them in a shortcut is feasible. Consecutive transit tasks passing through the robot's \home{} pose are merged as a single task. This allows the \home{} pose to be skipped. A shortcut path directly interpolates between $C_n^i$ and $C_{n'}^i$ to generate a sequence of poses with the same step size $\Delta t$. The shortcut must be collision-checked against any independent nodes (\ie nodes that are not predecessors to $v_n^i$ or successors to $v_{n'}^i$ in the TPG). On a multi-modal TPG, the collision checking must include any attached objects to the robot, as well as any independent object nodes on a task graph (\ie object nodes that are not predecessors or successors of the current task). Once a valid shortcut is found, the original nodes between $v_n^i$ and $v_{n'}^i$ are replaced by new nodes corresponding to the shortcut. If adding a valid shortcut from $v_n^i$ to $v_{n'}^i$ skips any outgoing edges between these two nodes, the start nodes of these outgoing edges are moved to $v_n^i$. If any incoming edges are skipped, then the end nodes of these incoming edges are moved to $v_{n'}^i$. These two steps ensure that any dependencies before adding a shortcut still exist after, and the TPG remains collision-free. Since each shortcut is collision-free, no new dependencies in the TPG are introduced, and TPG remains deadlock-free. The shortcutting algorithm keeps identifying valid shortcuts until a user-defined time limit is reached, removing redundant transitions to \home{} poses in the process.  

\subsection{TPG Execution}
\label{sec:tpg_exec}
Executing a motion plan on robot arms often requires a position controller for movement and other specific controllers for manipulation skills. These controllers may have delays or uncertainties that affect the real-robot execution time. The TPG formulation provides an easy way to execute a multi-robot plan. Here, we present a semi-centralized mechanism to coordinate multiple robot arms. 

As shown in \cref{fig:tpg execute}, the TPG is hosted on a central server that communicates with each robot's execution thread. Each pose node in the TPG corresponds to an action that moves the robot's position to the node's configuration. Each skill node corresponds to an action that executes the predefined robot skill. An action can be executed safely if there are no incoming edges from any nodes that are not executed. Although each TPG node is associated with a timestamp, it is only used for TPG construction and is ignored during execution.

If an action is safe to execute based on the TPG, the central server sends it to the robot's action queue. Each robot maintains its own controller-sensing loop and actuates the robot according to upcoming commands and its state estimation. The state estimation is also shared with the TPG, which then updates the TPG when a node is being executed or completed. Newly completed nodes can enable the central server to enqueue new nodes if their outgoing edges were previously preventing unsafe actions. During execution, TPG can be interpreted as a control law that maintains the safe scheduling of individual robot actions.

\begin{figure}[t]
\centering
\subfigure[]{\includegraphics[width=0.31\linewidth]{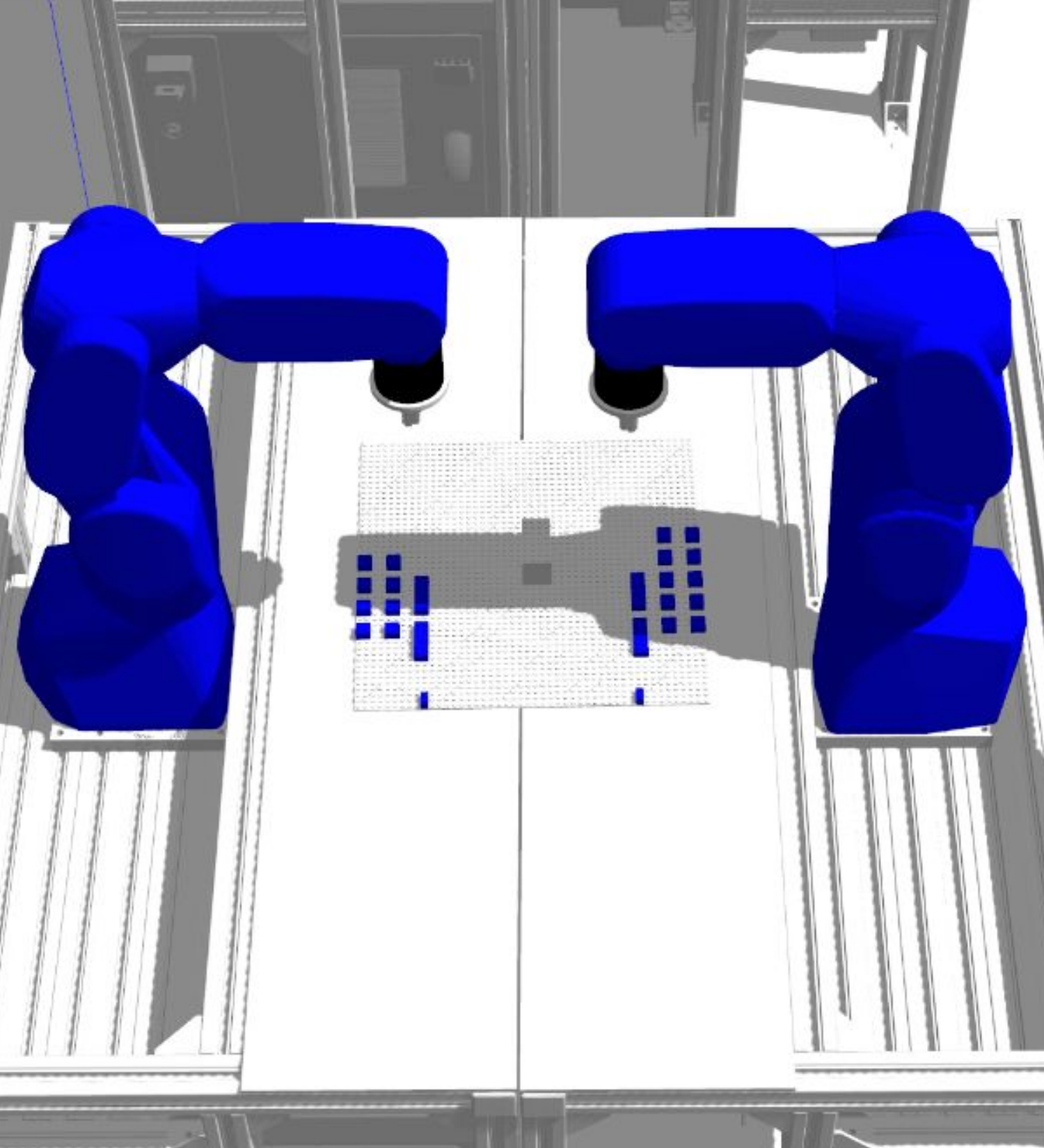}\label{fig:sim_env_setup}}\hfill
\subfigure[]{\includegraphics[width=0.31\linewidth]{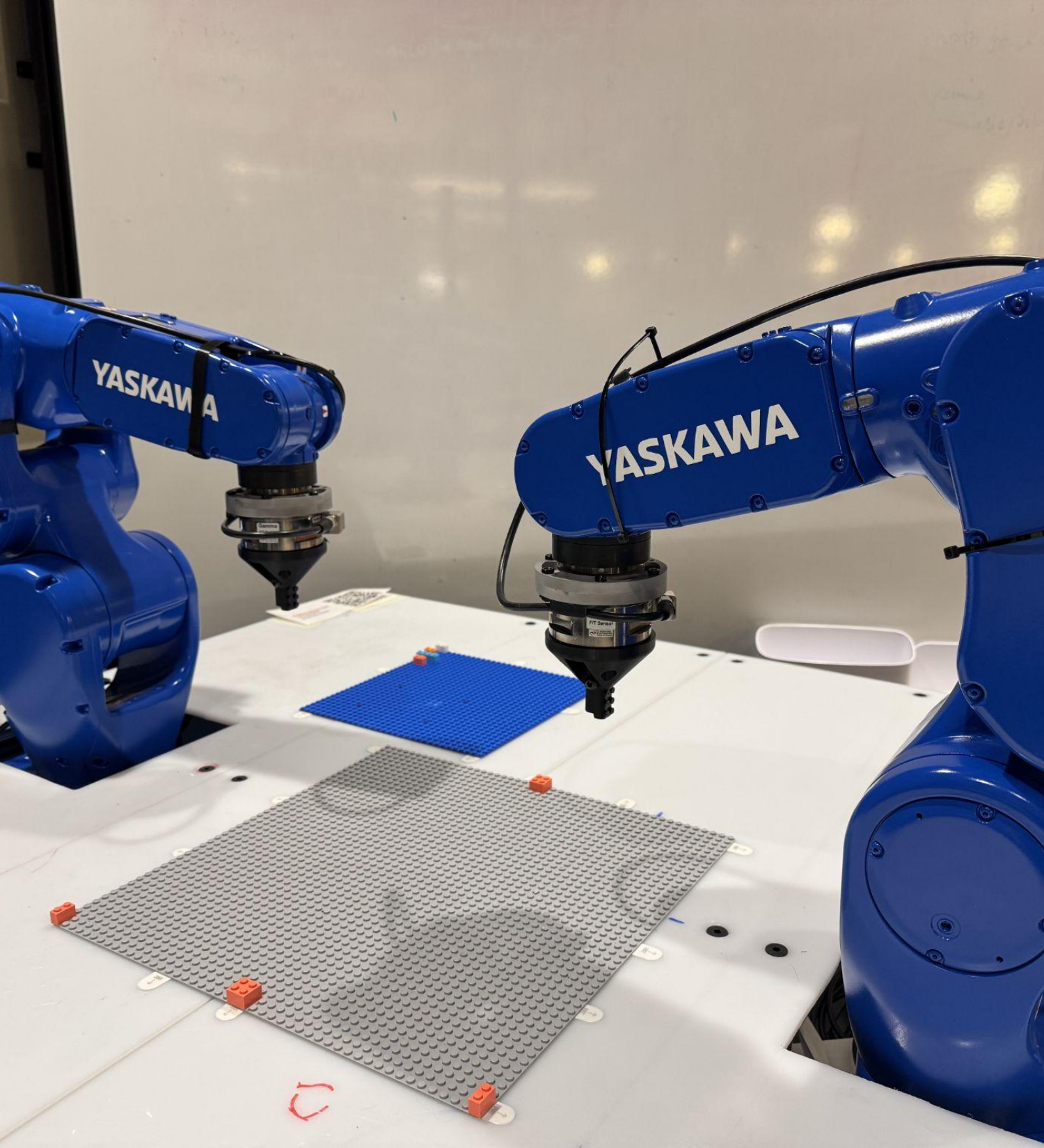}\label{fig:env_setup}}\hfill
\subfigure[]{\includegraphics[width=0.31\linewidth]{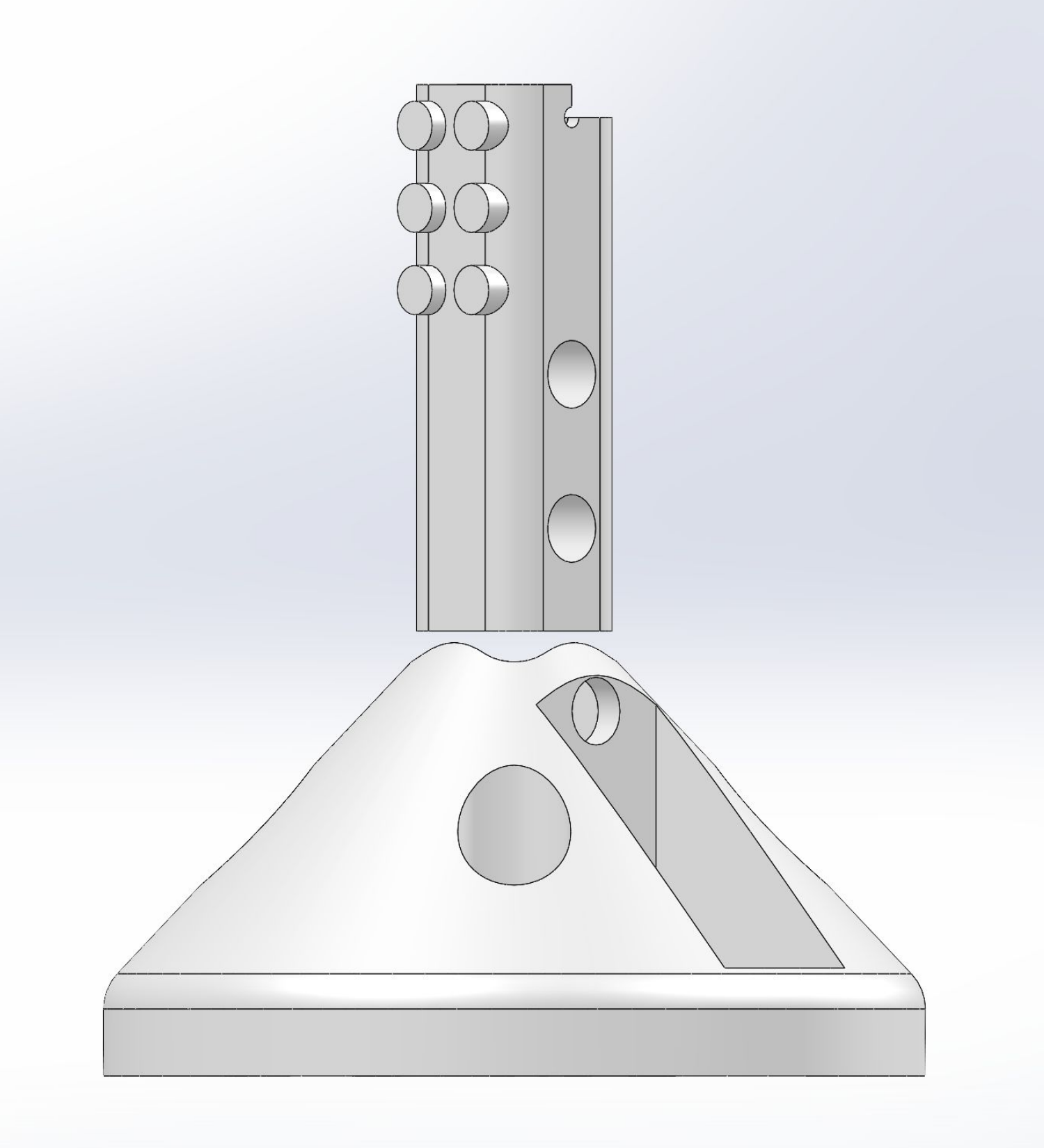}\label{fig:eoat}}
    \caption{Experiment setup for bimanual \LEGO{} assembly. (a) Simulation environment. (b) Real setup. (c) Illustration of the EOAT, \ie \tool{}, for the robot to construct \LEGO{} structures. \label{fig:exp_setup}}
    \vspace{-20pt}
\end{figure}

\section{Results}\label{sec:results}

\begin{figure}[t]
\centering
\subfigure[Cliff.]{\includegraphics[width=0.33\linewidth]{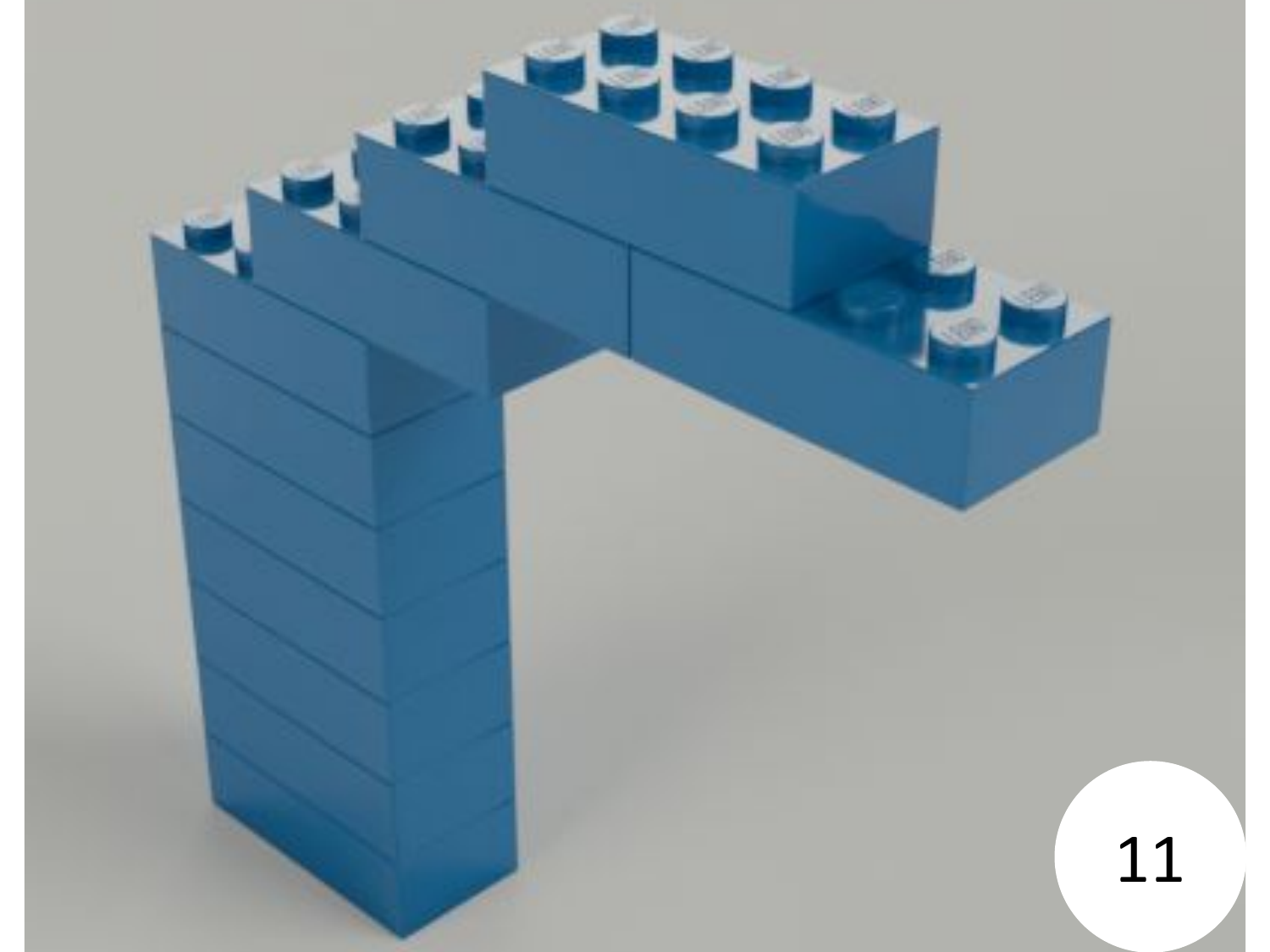}\label{fig:cliff_render}}\hfill
\subfigure[Branched stairs.]{\includegraphics[width=0.33\linewidth]{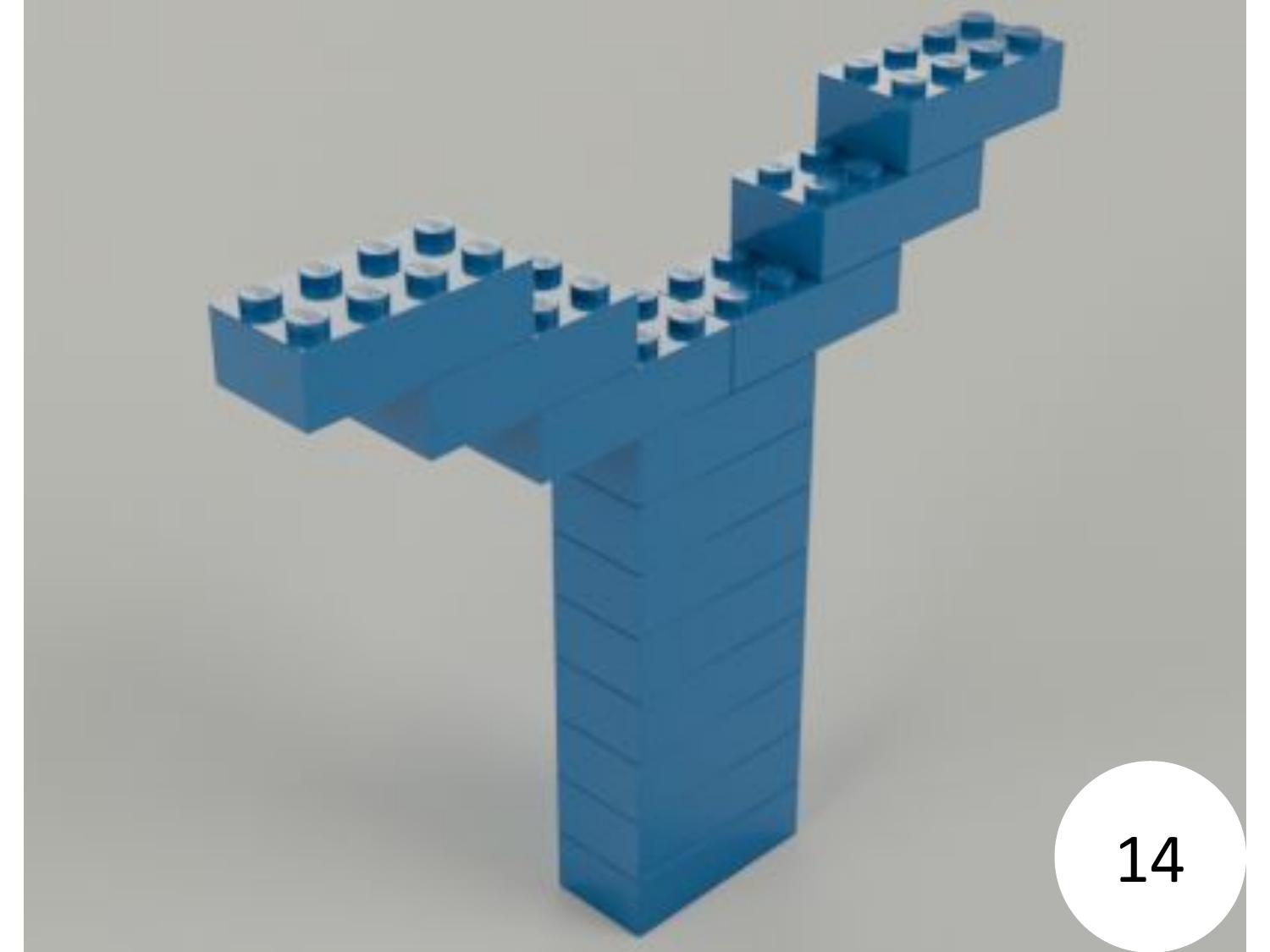}\label{fig:branched_stairs_render}}\hfill
\subfigure[Faucet.]{\includegraphics[width=0.33\linewidth]{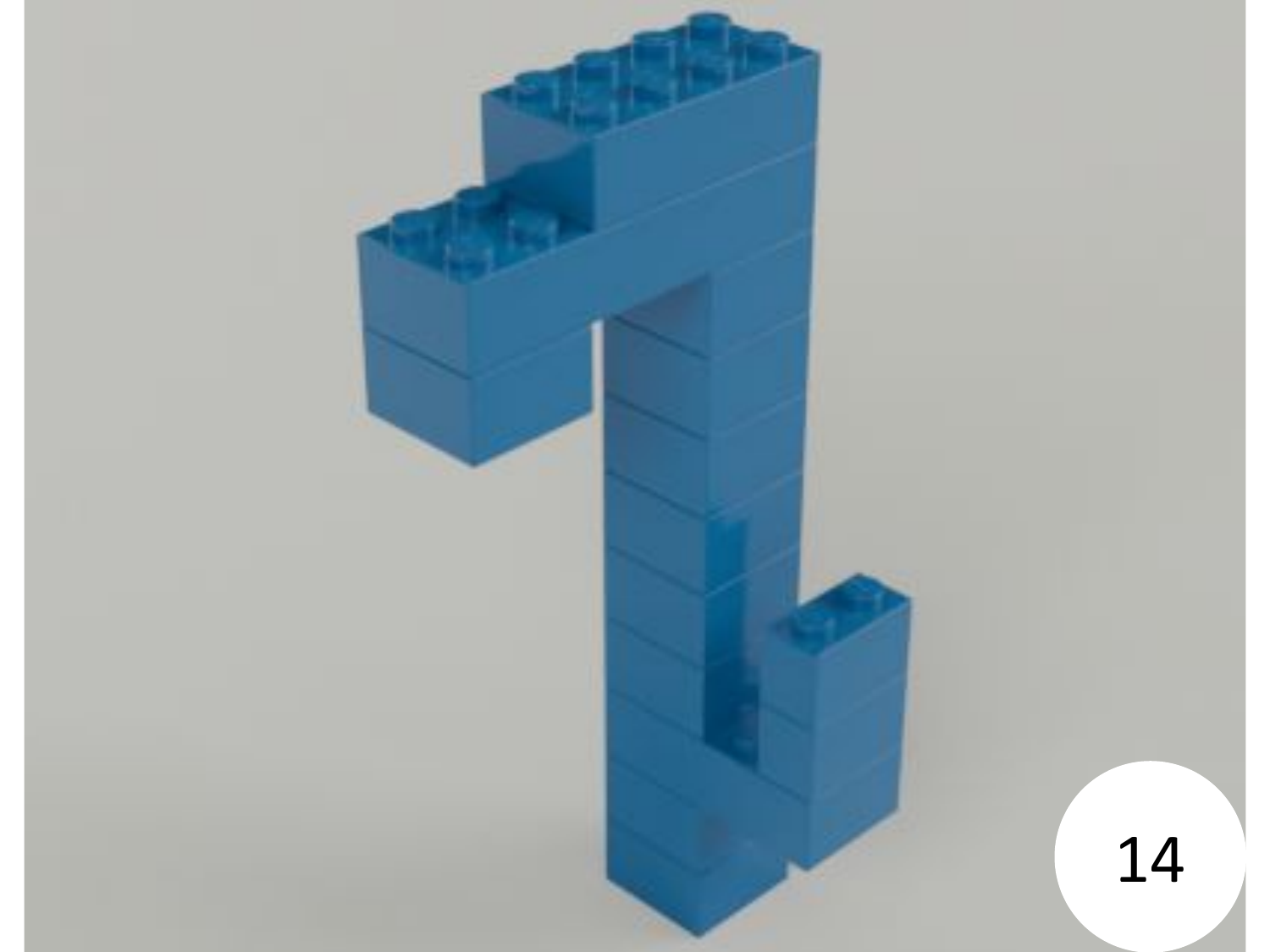}\label{fig:faucet_render}}\\
\subfigure[Bridge.]{\includegraphics[width=0.33\linewidth]{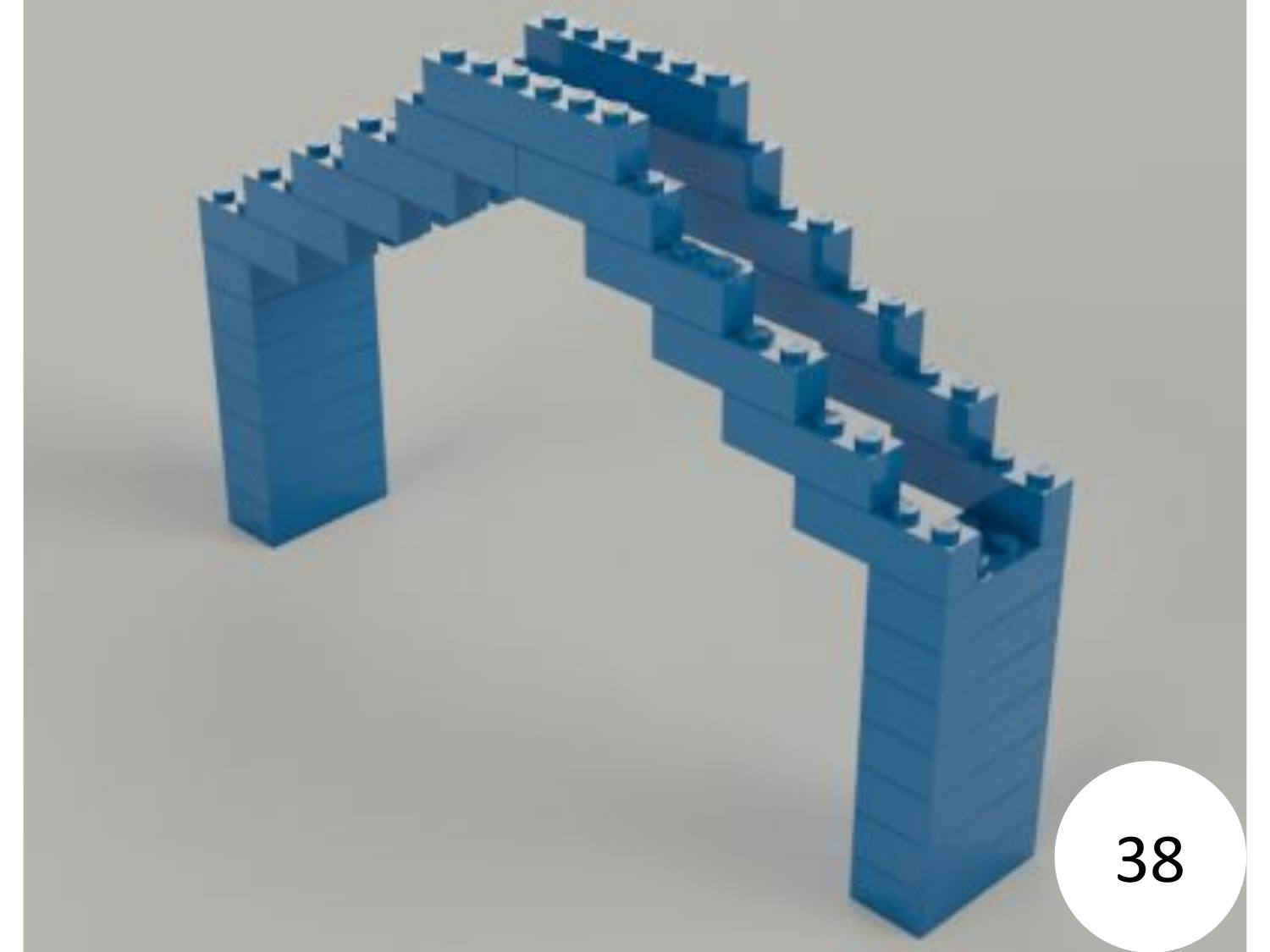}\label{fig:bridge_render}}\hfill
\subfigure[Fish.]{\includegraphics[width=0.33\linewidth]{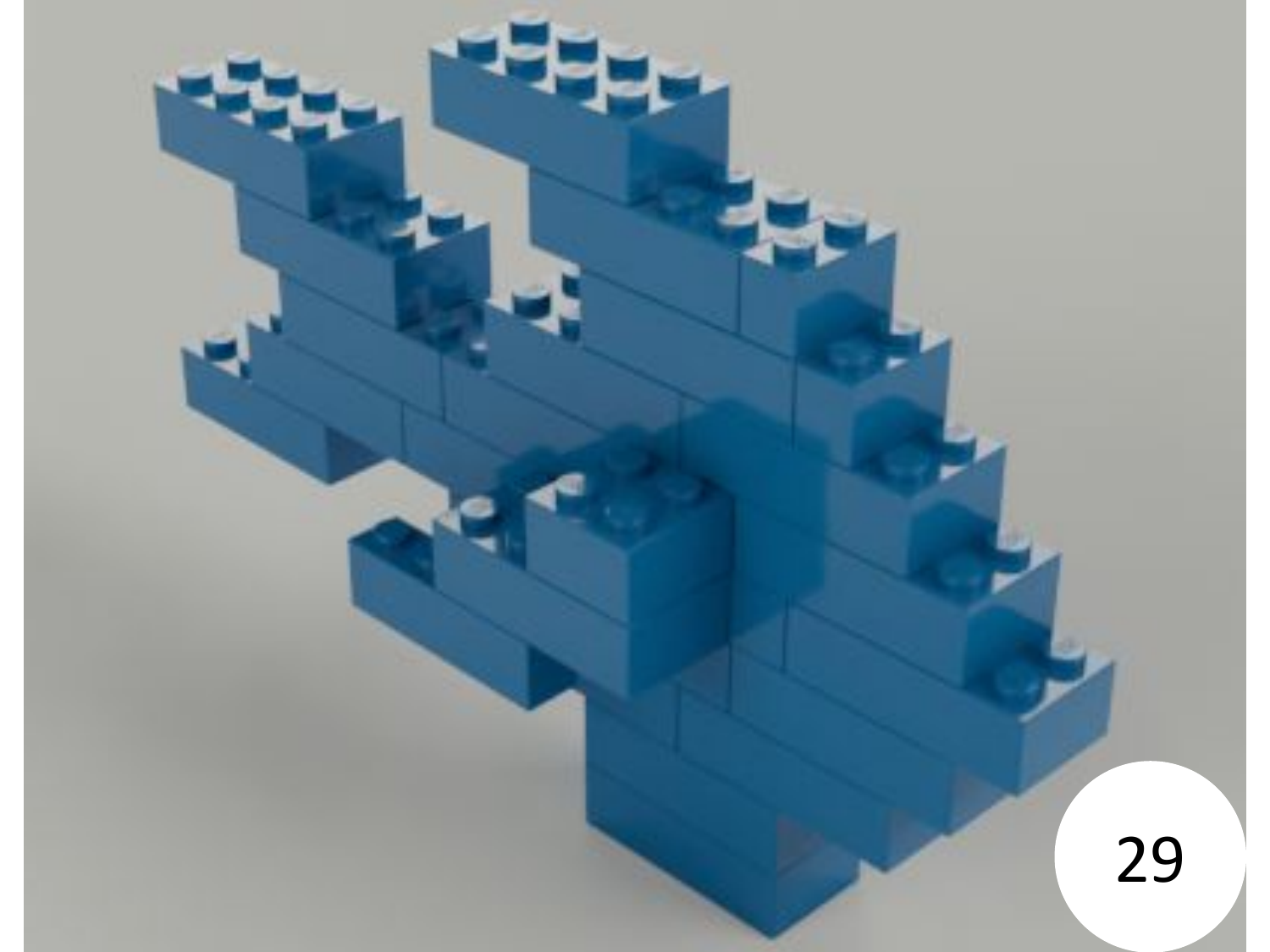}\label{fig:fish_render}}\hfill
\subfigure[Chair.]{\includegraphics[width=0.33\linewidth]{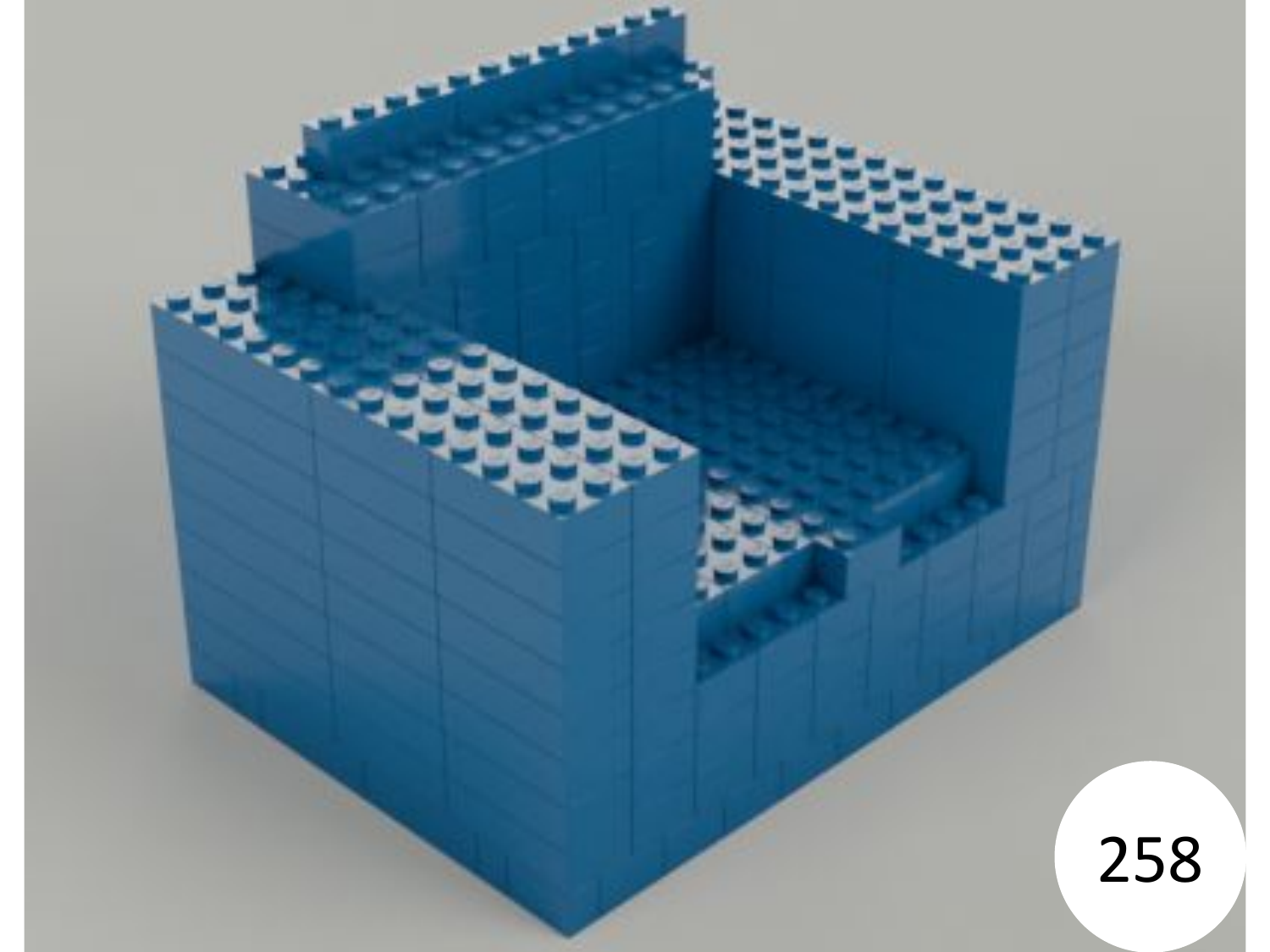}\label{fig:chair_render}}\\
\subfigure[Vessel.]{\includegraphics[width=0.33\linewidth]{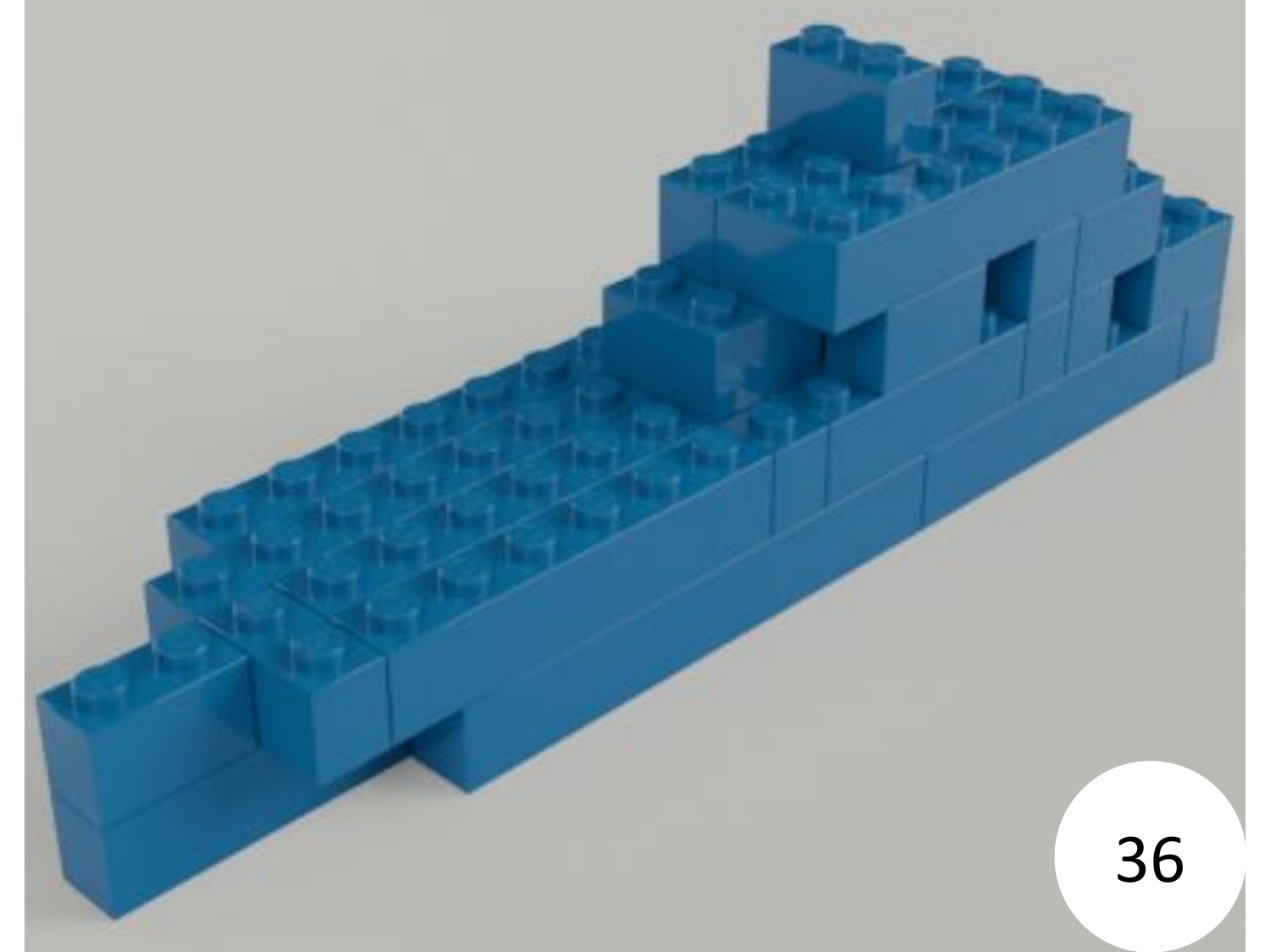}\label{fig:vessel_render}}\hfill
\subfigure[Guitar.]{\includegraphics[width=0.33\linewidth]{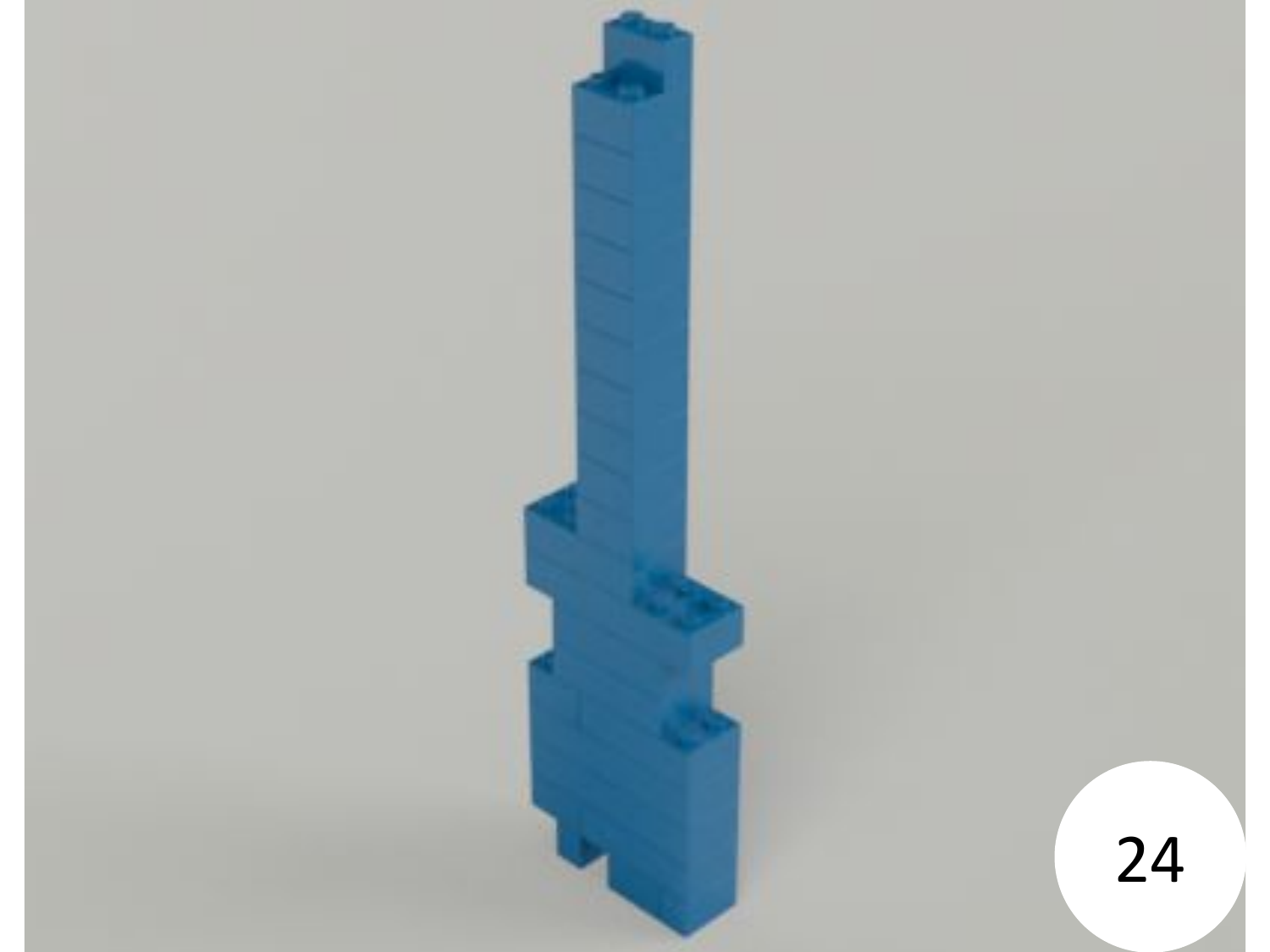}\label{fig:guitar_render}}\hfill
\subfigure[RSS.]{\includegraphics[width=0.33\linewidth]{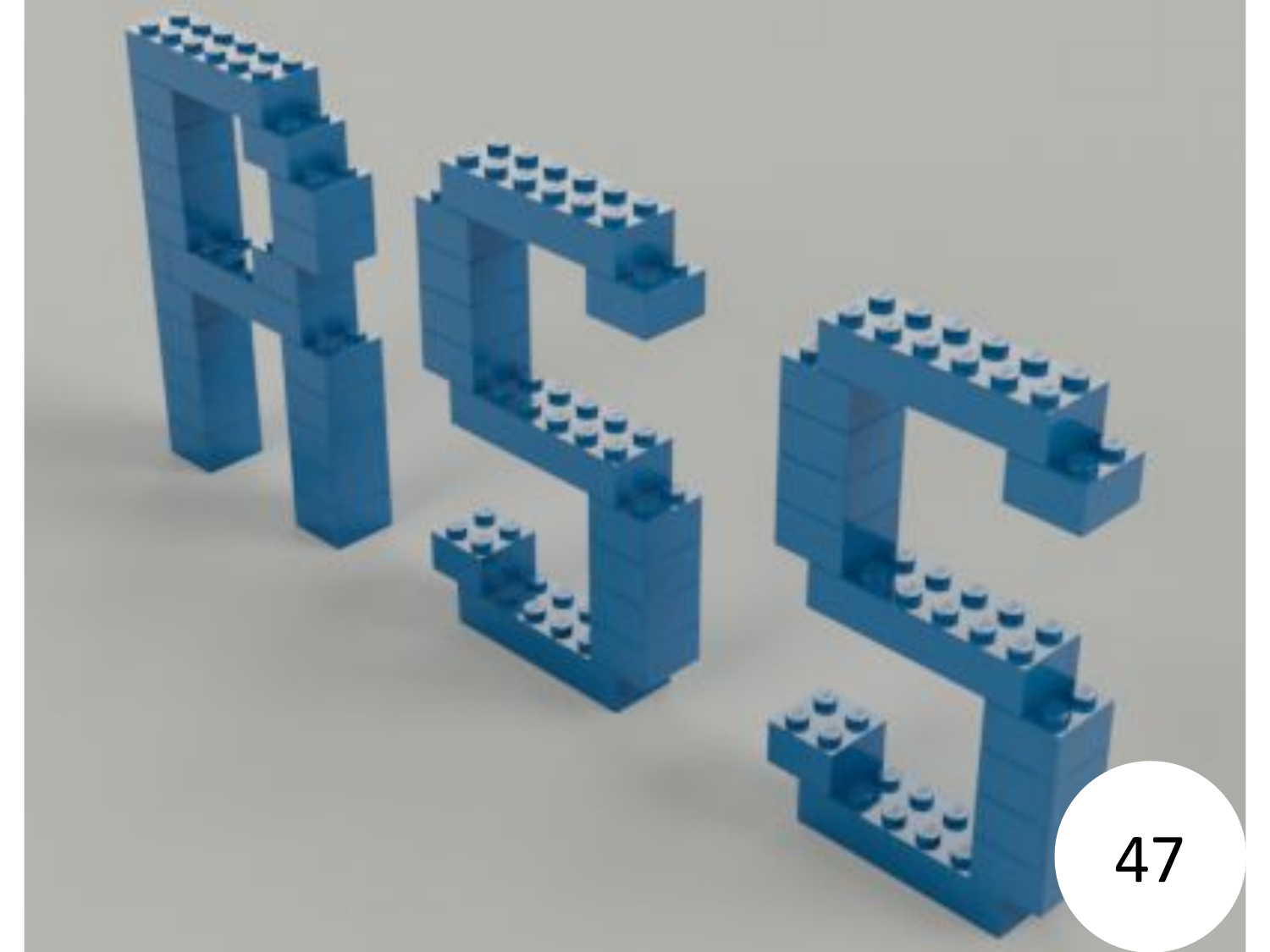}\label{fig:rss_render}}
    \caption{Customized \LEGO{} designs for evaluating \apex{}. The number in each figure indicates the number of objects required to assemble the \LEGO{} structure. \label{fig:eval_lego}}
    \vspace{-10pt}
\end{figure}

\begin{figure}
\centering
\subfigure[Cliff.]{\includegraphics[width=0.33\linewidth]{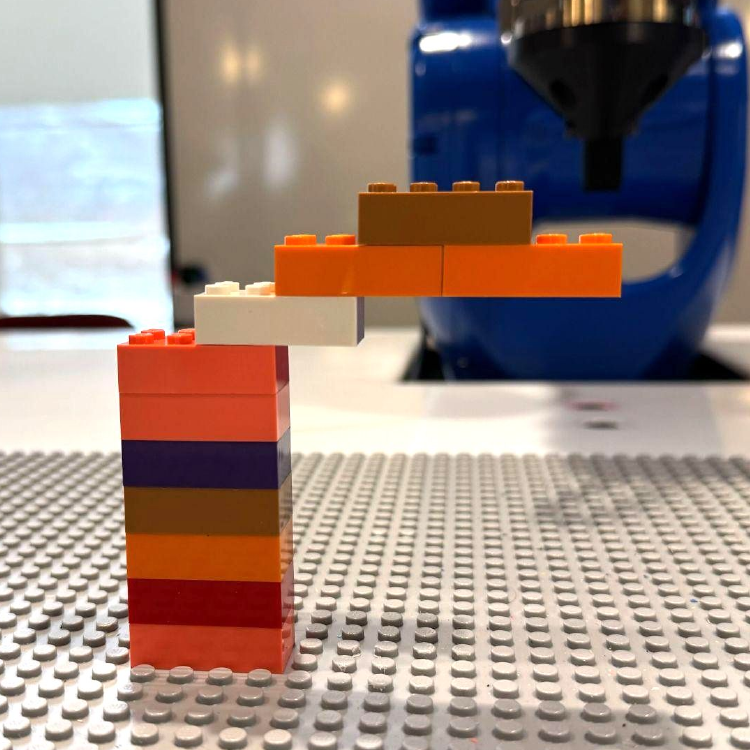}\label{fig:cliff}}\hfill
\subfigure[Branched stairs.]{\includegraphics[width=0.33\linewidth]{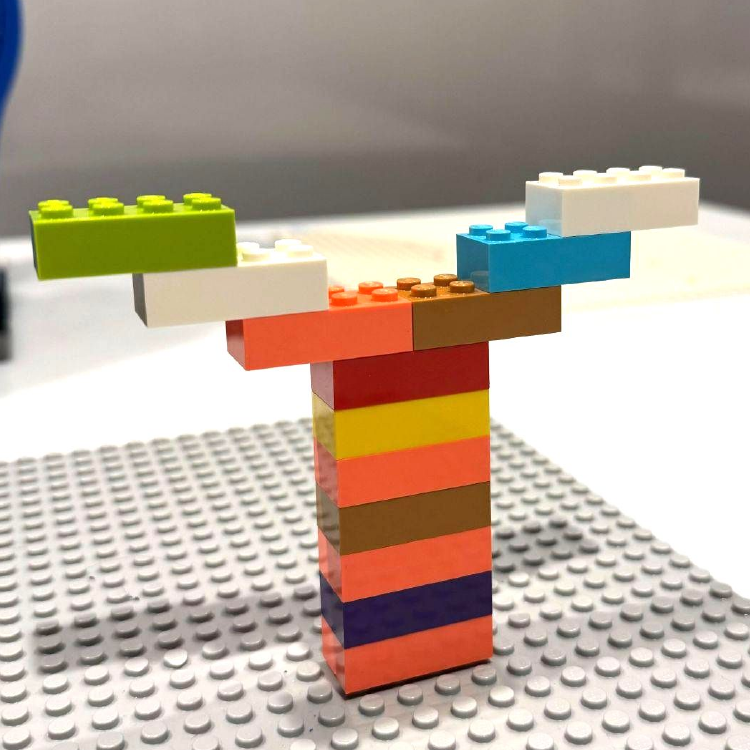}\label{fig:branched_stairs}}\hfill
\subfigure[Faucet.]{\includegraphics[width=0.33\linewidth]{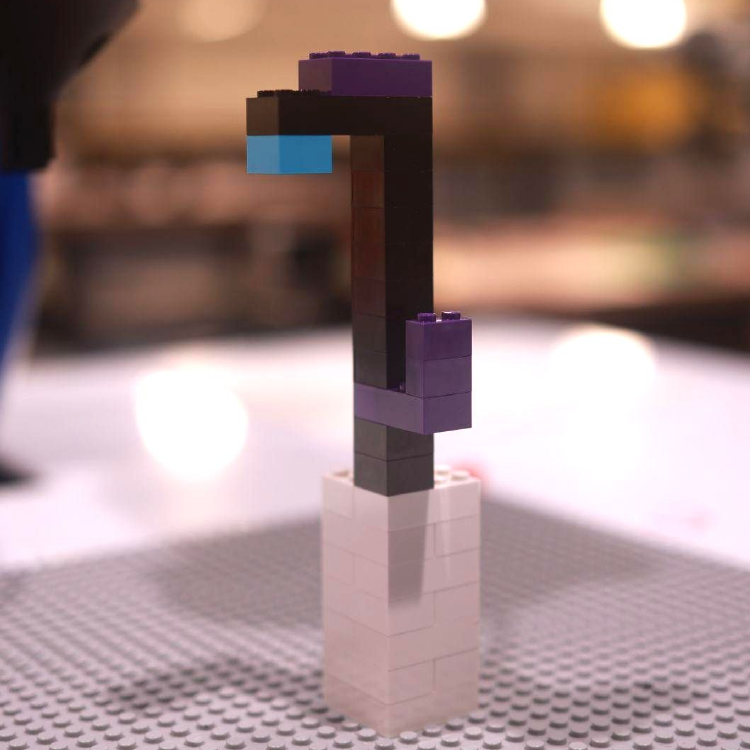}\label{fig:faucet}}\\
\subfigure[Vessel.]{\includegraphics[width=0.33\linewidth]{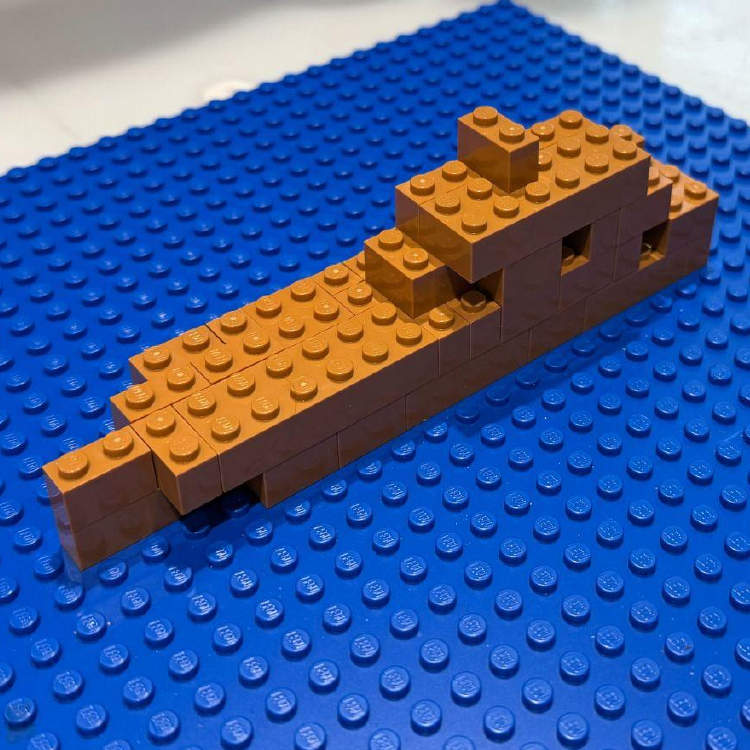}\label{fig:vessel}}\hfill
\subfigure[Guitar.]{\includegraphics[width=0.33\linewidth]{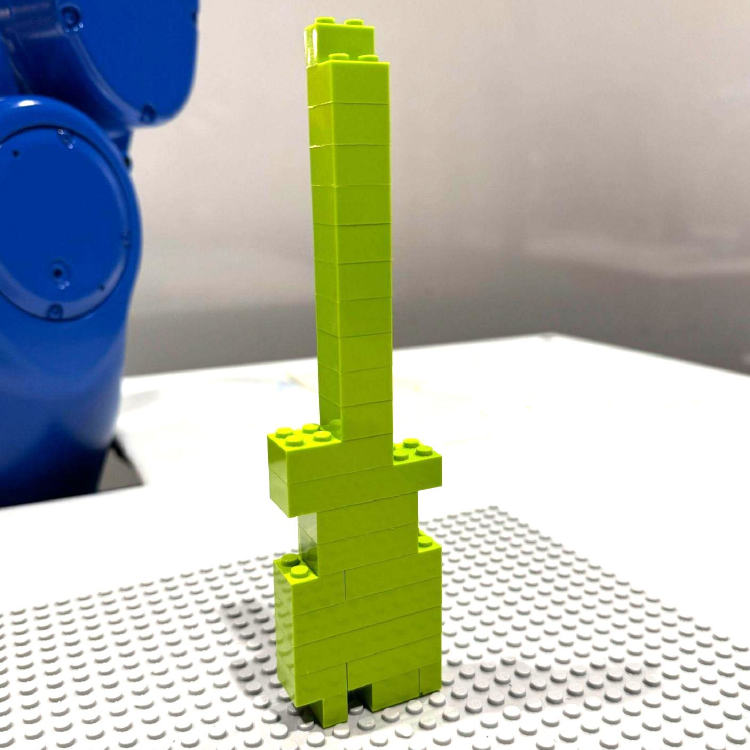}\label{fig:guitar}}\hfill
\subfigure[RSS.]{\includegraphics[width=0.33\linewidth]{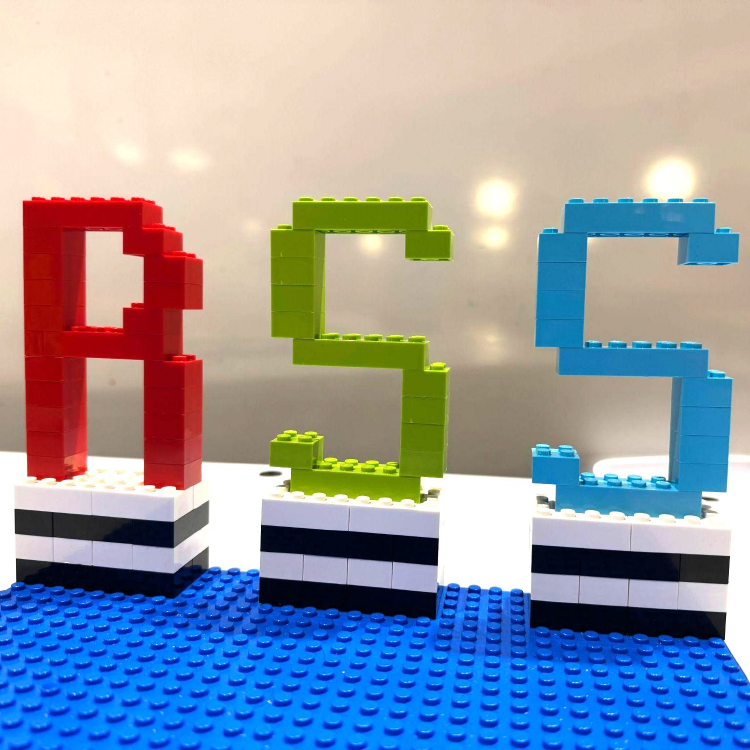}\label{fig:rss}}
    \caption{Example \LEGO{} structures constructed in real by the dual-arm system. \label{fig:lego_assembly_structures}}
    \vspace{-15pt}
\end{figure}

To evaluate performance, we apply the proposed \apex{} to bimanual \LEGO{} assembly tasks.
Given a customized \LEGO{} design as shown in \cref{fig:eval_lego}, \apex{} coordinates the robots to construct the desired structure as shown in \cref{fig:lego_assembly_structures} using available \LEGO{} bricks.
We deliberate the inputs (introduced in \cref{sec:prelim}) to \apex{} 
below. 

\textbf{Environment Setup}
\cref{fig:sim_env_setup,fig:env_setup} illustrate the simulation environment and the real setup, which includes two Yaskawa GP4 robots.
Following the task convention in \cite{Liu2023-ny}, we consider building \LEGO{} structures on a baseplate, which is calibrated\footnote{We calibrate the transformation from the robots to the baseplate by teleoperating the robot to touch the plate. Note that we only measure the translation ($X$, $Y$, $Z$) and yaw angle while assuming no roll and pitch offsets.} and placed between the two robots, using commercial standard \LEGO{} bricks initially stored on the baseplate.
Each robot is equipped with an ATI Gamma force-torque sensor (FTS), and the end-of-arm tool (EOAT) is mounted on the FTS.
The simulation consists of the entire workspace, which includes robots, FTS, EOAT, \LEGO{}s, nearby workstations, etc. 
The complete digital environment provides rich and accurate information for \apex{} to safely coordinate robot collaboration.

\textbf{Assembly Plan}
Given a \LEGO{} structure, we employ the physics-aware assembly planning in \cite{Liu2024-hk} with customized \LEGO{} physics reasoning \cite{Liu2024-go} to generate a physically valid assembly sequence.
Specifically, a physically valid assembly sequence 
enforces that for each step after assembling a brick, the structure is stable and does not collapse.
More details on generating the physically valid assembly sequence are provided in \cref{sec:app-asp}.
Note that the definition of a physically valid assembly sequence can be different for other cooperative assembly tasks.
For other applications, the assembly sequence can be obtained via planners, \eg \cite{10.1145/3550454.3555525}, and the proposed \apex{} is also applicable downstream. 

\textbf{Manipulation Skills}
Manipulating \LEGO{} bricks is a non-trivial contact-rich manipulation problem beyond simple pick and stack. 
A robot EOAT and manipulation policy (\ie insert-and-twist) were presented in \cite{Liu2023-ny}, which enable a robot to manipulate commercial standard \LEGO{} bricks, \ie pick, and place-down in \cref{fig:dual_arm_example}.
However, a robot can only use it to manipulate a \LEGO{} brick from its top, which limits the system from constructing complex structures.
To enhance the system capability, we present a new \LEGO{} tool (\tool{}), as shown in \cref{fig:eoat}.
In particular, \tool{} has \LEGO{} studs added to the side of the tooltip.
The new design enables the robot to manipulate a brick from its bottom as shown in \cref{fig:dual_arm_example}, \ie handover and place-up.
With \tool{}, we define the manipulation skills as shown in \cref{fig:dual_arm_example}, including 1) goal reaching with force feedback (\ie support-bottom and support-top), and 2) learned force policy (\ie pick, place-down, place-up, handover). 
More details on \LEGO{} manipulation skills are discussed in \cref{sec:app-skill}.

\begin{figure*}
    \centering
    \includegraphics[width=\linewidth]{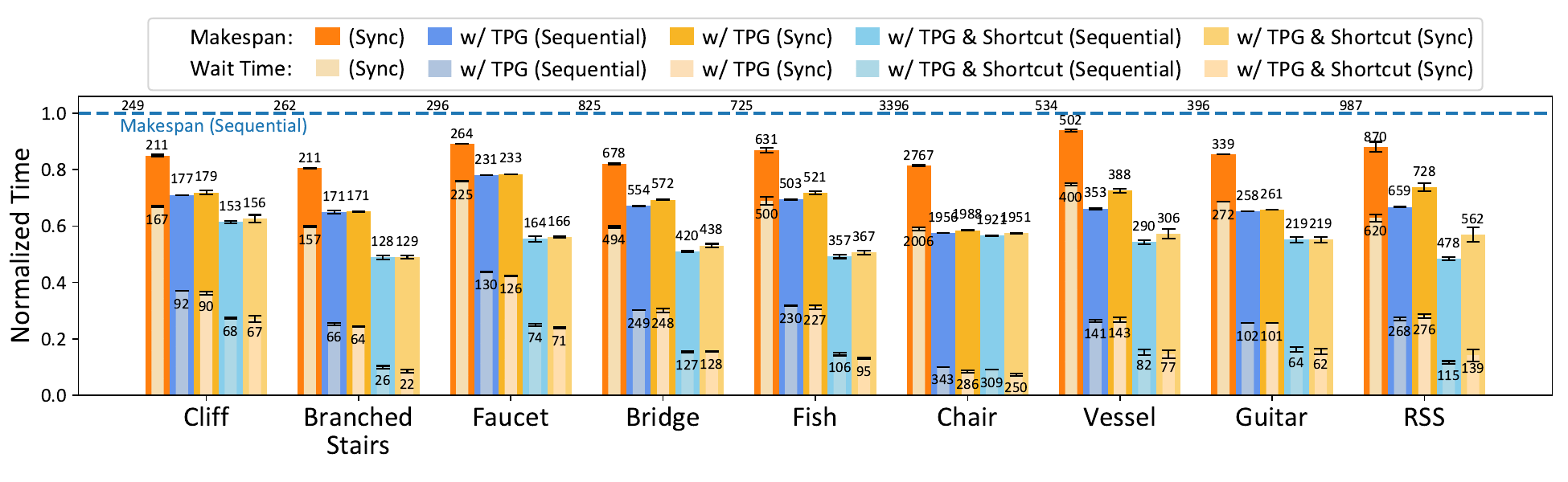}
    \caption{Normalized makespan and wait time of \apex{} (Sequential) versus the synchronized planner across example evaluation environment. The results are normalized by the makespan of the sequential motion plan before TPG and averaged over 4 random seeds. The unnormalized makespan and wait time in seconds are labeled for each entry, and the dashed horizontal line corresponds to the makespan of the sequential motion plan before TPG. Shortcut refers to the anytime shortcutting algorithm on TPG in \cref{sec:asynch_exec}.
    }
    \label{fig:makespan_waittime}
\end{figure*}

\begin{figure*}
    \centering
    \includegraphics[width=\linewidth]{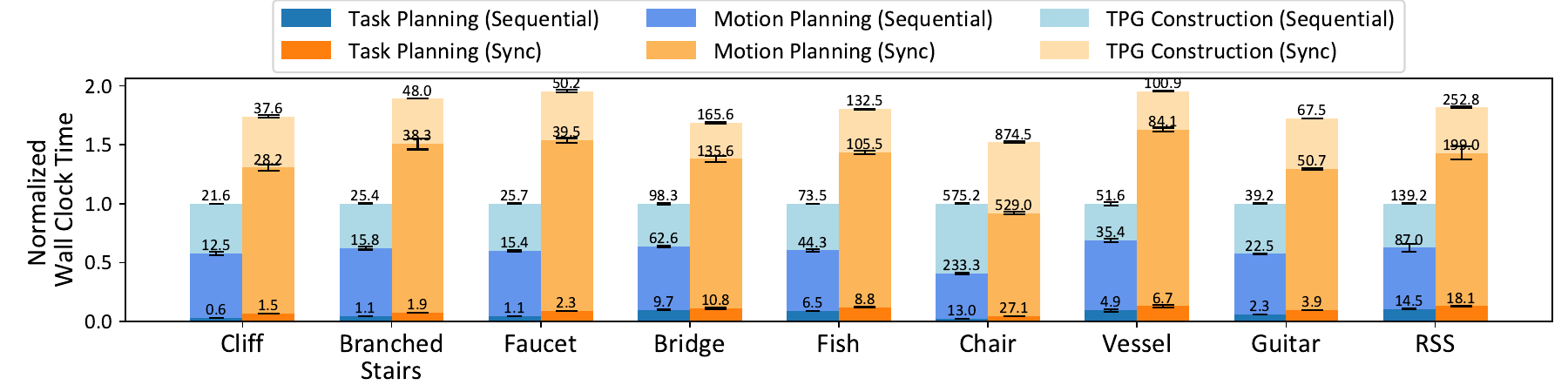}
    \caption{Breakdown of wall clock time of \apex{} (Sequential) and the synchronized planner baseline across evaluation environments. The results are normalized by the total of \apex{} planning time for each task (excluding shortcutting) and averaged over 4 random seeds. The running sum of unnormalized wall clock time in seconds is labeled for each component.  }
    \label{fig:time-plot}
\end{figure*}

\textbf{Implementation} We implement the TPG algorithm and manipulation skills in C++ with ROS-Noetic and \textit{MoveIt} \citep{coleman2014reducing}. The ILP in task planning is solved with the pulp Python package. The RRT-Connect motion planning uses \textit{MoveIt}'s OMPL \citep{OMPL2012} plug-in. Paths are discretized using $\Delta t = 0.05$ seconds when the maximum $L_1$ joint velocity is 1 rad/s. $\Delta t$ is adjusted linearly based on the maximum velocity to ensure the same density. All simulation experiments are conducted on an AMD 7840HS laptop.
Algo. \ref{algo:mmtpg} parallelizes collision checking with 16 threads.

\textbf{Experiment Objective}
While \apex{} itself is a full pipeline for multi-robot tasks and motion planning, the key innovation that enables asynchronous collaboration is the TPG execution framework. Thus, we are interested in the following questions when evaluating \apex{}:
\begin{itemize}
    \item (Q1) How significant is the benefit of asynchronous execution, enabled by the TPG execution framework?
    \item (Q2) How is the quality of plans produced by \apex{} and what are the computational costs?
    \item (Q3) How well does \apex{} perform in physical \LEGO{} assembly, and can it safely execute planned paths despite uncertainties?
\end{itemize}

Q1 and Q2 will be closely examined in simulation, whereas Q3 will be the focus of our real robot experiments.

\subsection{Simulation Performance} 

We first conduct experiments in simulation. To our knowledge, no existing simulator can reliably simulate the connections between \LEGO{} bricks. Thus, all robot skills are reduced to deterministic operations when evaluated in simulation, and any variations are due to the stochasticity of planning in \apex{}. 

\textbf{Dataset} We evaluate the performance of \apex{} on a suite of nine \LEGO{} assembly tasks as shown in \cref{fig:eval_lego}. The complexity of these tasks varies significantly in terms of the number of objects in the assembly plan, stability, orientation, and manipulation skills required for physical assembly. 
The Chair shown in \cref{fig:chair_render} has 258 objects, but the structure is solid and stable, and thus, no collaborative skills are needed. 
On the other hand, many of the bricks along the span of the Bridge (\cref{fig:bridge_render}) require a robot to support them when assembling from the top, whereas building the Cliff (\cref{fig:cliff_render}) and Faucet (\cref{fig:faucet_render}) requires an object reorientation and collaborative assembly from bottom.

\textbf{Metrics} We use execution makespan and wait time as our evaluation metrics for plan quality. 
Since our output is a TPG, we first roll out the asynchronous path from the TPG, assuming no controller delay. 
The rollout path converts each pose node back to a configuration.
Actions in the skill nodes are executed based on the reference path, and force feedback is switched off in the simulation.
The timestamp for each configuration in the rollout path is the earliest possible time to reach this node based on incoming type-2 edges. 
Execution makespan is the maximum time taken among all robots to execute their rollout paths, \ie $\max_{i=1}^N t^i_{N^i_{end}}$. 
Wait time is defined as the total amount of time any robot spends waiting in the rollout asynchronous path from TPG, or the original sequential or synchronous path.
In particular, we are interested in whether TPG processing can successfully reduce wait time when initialized with a sequential task and motion plan.

\textbf{Baseline} We also design a baseline for synchronized task and motion planning as a comparison to \apex{}. 
Although \apex{} uses a sequential planner for simplicity and efficiency, our TPG can also improve synchronized motion plans, common in MR-TAMP. Thus, we evaluate the performance improvement of TPG on synchronized plans.
In this synchronous planner, robots can execute tasks in parallel but must wait for all robots to finish their current task before proceeding to the next set of tasks. 
We use an algorithm to convert the sequential task plan from \apex{} to a synchronous task plan, as shown by \cref{fig:seq_v_sync} in \cref{sec:app-seq_v_sync}.
The main idea is to execute the sequential task plan in parallel if executing the next task does not violate inter-robot task dependencies or block tasks scheduled at an earlier time. 
This process is similar to building a TPG on tasks instead of motions for parallelization.
For every robot $i$ and its task $\mathcal{T}_m^i$, the algorithm checks if the intermediate goal pose $C^i_{N_{end, m}^i}$ collides with any other intermediate goal pose of robot $i'$ and task $\mathcal{T}_{m'}^{i'}$ that satisfies $m' < m$. If there exists a collision, then robot $i$ must wait for robot $i'$ to complete task $\mathcal{T}_{m'}^{i'}$ before starting task $\mathcal{T}_m^i$. A synchronous task graph can then be generated by combining these calculated dependencies with existing inter-robot dependencies. 
Then, composite RRT-Connect is used as the multi-robot motion planner. Synchronous paths for tasks executing in parallel are generated by planning all degrees of freedom as a single robot.

\textbf{Performance} \cref{fig:makespan_waittime} shows the quality of the solution of \apex{} on a variety of tasks. 
First, the TPG post-processing and applying and shortcut, as described in \cref{sec:asynch_exec}, significantly reduces makespan by 48\% and wait time by 85\% on average, compared to the initial sequential motion plan on the horizontal dashed line. 
Compared to the synchronized motion plan, our asynchronous plans from \apex{} are consistently shorter and have lower wait time. 
When applied to the synchronous plan, TPG also significantly reduces the makespan by 36\% and wait time by 77\% on average. 
Note that the post-processed sequential plan from \apex{} still slightly outperforms the post-processed synchronized motion plan by 3\% in terms of makespan. 
This is due to the path produced by a multi-robot motion plan being suboptimal compared to sequential motion planning. 
Still, the wait time for the synchronized plan after TPG post-processing is minimal. 

\textbf{Runtime} \cref{fig:time-plot} and Table \ref{tab:planning_metrics} in \cref{sec:app-metrics} shows the wall clock time for \apex{}. 
On average, the TPG construction time is always lower than the task motion planning time except for the Chair, which has a very long assembly sequence. 
On the other hand, running a synchronized planner can be much more expensive than the simple sequential planner used in \apex{}, which requires more careful coordination in task planning and multi-robot motion planning with more degrees of freedom. 
By combining a simple sequential task and motion planner with TPG post-processing, \apex{} produces higher-quality multi-robot plans with 26\% lower computational overhead on average than a synchronized multi-robot task and motion planner alone.
One concern is that the number of collision checks when building TPG scales quadratically with the number of robots. Thus, we perform additional experiments in \cref{sec:app-exp} with three and four robot arms to show that the runtime of TPG construction remains reasonable.

\subsection{System Deployment}

\begin{figure}
    \centering
    \includegraphics[width=\linewidth]{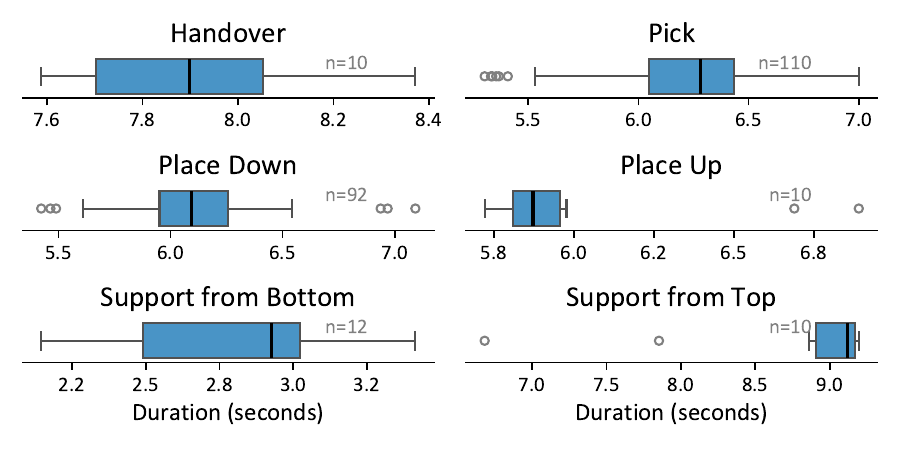}
    \caption{Distribution of executing various FTS-feedback-controlled \LEGO{} manipulation policy on real robot. The execution time is collected from assembling the Cliff structure in \cref{fig:cliff} repeatedly. This variation adds to the uncertainty in real-time execution.}
    \label{fig:task_duration}
    \vspace{-10pt}
\end{figure}

We deploy the proposed \apex{} to a real bimanual setup for cooperative \LEGO{} assembly.
\cref{fig:env_setup} illustrates the environment setup of the dual-arm system. 
Note that despite the environment being pre-calibrated, errors ($\sim 1$mm) still exist since the calibration is imperfect and the structure could be tilted due to the passive connection nature. 
Thus, we integrate real-time force feedback using the FTS to improve the manipulation robustness.
For operation skills (\ie pick, place-down, place-up, handover), we use the force feedback to detect successful insertion and update the manipulation policy accordingly.
For supporting skills (\ie support-bottom and support-top), we use force feedback to sense a slight touch with the structure to avoid either over or under-supporting.
\cref{fig:lego_assembly_structures} showcases example \LEGO{} structures accomplished by the bimanual system with \apex{}.
The robots can safely collaborate and efficiently build customized and complex \LEGO{} objects, including fragile overhanging structures.

Note that the key difference between assembling in real and in simulation is with and without force feedback.
Integrating force feedback into the manipulation skills improves the system robustness, but also brings uncertainty with respect to execution time.
In particular, we are interested in whether the TPG execution framework ensures safe execution and avoids collisions despite uncertainties of manipulation skills. 
\cref{fig:task_duration} depicts the distribution of execution times of six manipulation skills, \ie pick, place-down, place-up, handover, support-bottom, and support-top (see \cref{fig:dual_arm_example} for illustrations).
All of these skills are designed to execute with force feedback to ensure proper contact between the EOAT and the object being manipulated and minimize the effect of imperfect calibration or a tilted structure. 
As a consequence of these mm-level adjustments with force feedback, the execution time can vary as much as 2 seconds, or as much as 23\% from the median. 
Nevertheless, our TPG execution framework can reliably adjust the robot schedule if any delay could cause a collision or require another robot to wait longer until a skill is completed.

In practice, \apex{} also allows the two robot arms to operate in close proximity asynchronously thanks to the use of TPG. 
For example, if one robot is stopped due to a controller issue or because the user presses emergency stop, the other robot would automatically stop if it is unsafe to continue its execution. 
Although each robot's feedback controller operates independently, only those actions that are deemed safe are passed to the robot's action queue. 
Thus, \apex{} can significantly reduce the risk of unsafe action in real multi-robot execution.

\section{Limitations and Future Work}\label{sec:discussion}

Although the proposed \apex{} pipeline enables efficient and safe execution for multiple robot arms, it still has several algorithmic limitations which we discuss below.

\textbf{Offline Computation} Currently, both the TPG processing and motion planning in \apex{} are performed offline before real execution. This can be a drawback in real assembly tasks, where new tasks are continuously assigned once the robots finish an existing assembly on an assembly line. Another limitation of offline computation is that the \apex{} cannot easily adapt to changes in the collision environment or assembly steps. A principled framework to address these lifelong planning techniques is to use a windowed multi-robot planner and only convert the first $n$ steps of the robot plan to a TPG, similar to how \citet{Varambally2022-ux} address mobile robot coordination in automated warehousing. Taking a reduced-horizon approach will significantly reduce planning time and allow plans to be continuously updated concurrently with execution.

\textbf{Planning for Robot Dynamics} While \apex{} are reliable and safe on real robots, \apex{} requires a good position or force controller for the robot because the generated plan does not consider robot dynamics, such as acceleration and jerk constraints. This is challenging because the planner must generate continuous velocity and acceleration profiles while also avoiding inter-robot collisions at all times. We plan to incorporate dynamics as part of the TPG post-processing step, such as solving a linear program on top of TPG-imposed constraints and velocity constraints, as suggested in \citet{Honig2016-ts}. Another interesting problem with robot arms is that speed constraints may be imposed on the task space, due to the attached object at the end-effector or even closed kinematics chains formed by concurrent manipulation \cite{Xian2017-fz}.

\textbf{Manipulation Policy} While the proposed \apex{} enables efficient and safe dual-arm cooperative \LEGO{} assembly, the current system assembles each object based on predefined skills.
With the additional force feedback, each single operation can be performed robustly.
However, due to the passive connection nature of \LEGO{} structures, \ie established connections could be gradually loosened and the structure can be tilted or even collapse due to subsequent operations, the long-horizon assembly could still fail. 
Therefore, the dual-arm system, at its current stage, is not robust enough to construct large-scale \LEGO{} structures that have multiple fragile overhanging geometries.
To further improve the robustness from a system perspective, our aim is to investigate methods for failure detection \cite{chen2024automating} and recovery, \eg reinforcing the connections that are loosened due to later operations.
Failure detection and recovery can also be integrated as part of an online replanning framework that dynamically reschedules the robot's tasks if a failure occurs and intervention becomes necessary.   

\textbf{Other Cooperative Assemblies} Building on the APEX-MR pipeline, we plan to extend its application to a broader range of cooperative tasks, such as the NIST Box Assembly \cite{nist-box}, and other industrial assembly scenarios. In doing so, we aim to address the unique challenges posed by real-world manufacturing environments, gaining deeper insights into how cooperative systems can be optimized for complex, large-scale production tasks. 
While the discussion in this paper is based on \LEGO{} assembly, the components in \apex{}, especially TPG post-processing, can be applied to other multi-robot assembly tasks. A concrete step would be to investigate how to integrate manipulation policies that are more complex than those used for \LEGO{} assembly, \eg diffusion policy \citep{chi2024diffusionpolicy}, to the TPG framework.


\section{Conclusion} \label{sec:conclusion}
\label{sec:conclusion}

For many robotic manipulation tasks, a team of cooperative robot arms is often necessary and beneficial because cooperation can improve dexterity, flexibility, and versatility. A reliable framework for coordinating robot arms should possess several key qualities: efficiency to maximize throughput, scalability to long-horizon and complex tasks, and safety during real execution. 

With these criteria in mind, we have proposed \apex{}, a pipeline for multi-robot asynchronous planning and execution. Our proposed pipeline combines a sequential task and motion planner with a TPG to post-process the plan for asynchronous execution. Specifically, TPG post-processing can significantly speed up the execution of otherwise sequential and synchronous multi-robot task plans by 48\% and 36\% on our simulated assembly tasks. Because coordination is easier, sequential motion planning is far more efficient than planning for synchronous execution.

We demonstrated that our proposed algorithm can be successfully deployed and integrated for a real bimanual cooperative task. LEGO structures, as an example, presented a challenging manipulation task due to the need for high precision and the non-rigid nature of their connections.
We presented a set of manipulation skills for complex cooperative assembly, including supported placement and object handover, based on the end-effector design LT-V2. The dual-arm system successfully performs customized \LEGO{} assembly and is the \textit{first} robotic system to do so with commercial \LEGO{} bricks. Additionally, we showed that the integration with the TPG execution framework is robust in the presence of uncertain execution time. In the end, we hope that this framework can advance and bring closer to more real use of multi-robot arm collaboration algorithms.

\section*{Acknowledgments}
This work is in part supported by the National Science Foundation (NSF) under grant number 2328671 and the Manufacturing Futures Institute, Carnegie Mellon University, through a grant from the Richard King Mellon Foundation, as well as a gift from Amazon.
The authors also thank Yifan Sun for helping design \tool{}.

\bibliographystyle{plainnat}
\bibliography{references}

\begin{thebibliography}{61}
\providecommand{\natexlab}[1]{#1}
\providecommand{\url}[1]{\texttt{#1}}
\expandafter\ifx\csname urlstyle\endcsname\relax
  \providecommand{\doi}[1]{doi: #1}\else
  \providecommand{\doi}{doi: \begingroup \urlstyle{rm}\Url}\fi

\bibitem[Berndt et~al.(2020)Berndt, Van~Duijkeren, Palmieri, and Keviczky]{Berndt2020-xq}
Alexander Berndt, Niels Van~Duijkeren, Luigi Palmieri, and Tamas Keviczky.
\newblock A feedback scheme to reorder a multi-agent execution schedule by persistently optimizing a switchable action dependency graph.
\newblock \emph{arXiv 2010.05254}, October 2020.

\bibitem[Bien and Lee(1992)]{Bien1992-ze}
Z~Bien and J~Lee.
\newblock A minimum-time trajectory planning method for two robots.
\newblock \emph{IEEE Transactions on Robotics and Automation}, 8\penalty0 (3):\penalty0 414--418, June 1992.

\bibitem[Brunello et~al.(2025)Brunello, Fabris, Gasparetto, Montanari, Saccomanno, and Scalera]{BRUNELLO2025102920}
Andrea Brunello, Giuliano Fabris, Alessandro Gasparetto, Angelo Montanari, Nicola Saccomanno, and Lorenzo Scalera.
\newblock A survey on recent trends in robotics and artificial intelligence in the furniture industry.
\newblock \emph{Robotics and Computer-Integrated Manufacturing}, 93:\penalty0 102920, 2025.

\bibitem[Chen et~al.(2024)Chen, Yao, Liu, Liu, and Ichnowski]{chen2024automating}
Hongyi Chen, Yunchao Yao, Ruixuan Liu, Changliu Liu, and Jeffrey Ichnowski.
\newblock Automating robot failure recovery using vision-language models with optimized prompts.
\newblock \emph{arXiv:2409.03966}, 2024.

\bibitem[Chen et~al.(2022)Chen, Li, Huang, Garrett, Sun, Fan, Hofmann, Mueller, Koenig, and Williams]{chen2022cooperativeMRAMP}
Jingkai Chen, Jiaoyang Li, Yijiang Huang, Caelan Garrett, Dawei Sun, Chuchu Fan, Andreas Hofmann, Caitlin Mueller, Sven Koenig, and Brian~C Williams.
\newblock Cooperative task and motion planning for multi-arm assembly systems.
\newblock \emph{arXiv:2203.02475}, 2022.

\bibitem[Chi et~al.(2024)Chi, Xu, Feng, Cousineau, Du, Burchfiel, Tedrake, and Song]{chi2024diffusionpolicy}
Cheng Chi, Zhenjia Xu, Siyuan Feng, Eric Cousineau, Yilun Du, Benjamin Burchfiel, Russ Tedrake, and Shuran Song.
\newblock Diffusion policy: Visuomotor policy learning via action diffusion.
\newblock \emph{The International Journal of Robotics Research}, 2024.

\bibitem[Coleman et~al.(2014)Coleman, Sucan, Chitta, and Correll]{coleman2014reducing}
David Coleman, Ioan Sucan, Sachin Chitta, and Nikolaus Correll.
\newblock Reducing the barrier to entry of complex robotic software: a \textit{MoveIt} case study.
\newblock \emph{arXiv:1404.3785}, 2014.

\bibitem[Erdmann and Lozano-Pérez(1987)]{Erdmann1987-hu}
Michael Erdmann and Tomás Lozano-Pérez.
\newblock On multiple moving objects.
\newblock \emph{Algorithmica}, 2\penalty0 (1-4):\penalty0 477--521, November 1987.

\bibitem[Fan et~al.(2019)Fan, Luo, and Tomizuka]{10.1109/ICRA.2019.8793659}
Yongxiang Fan, Jieliang Luo, and Masayoshi Tomizuka.
\newblock A learning framework for high precision industrial assembly.
\newblock In \emph{Proceedings of the IEEE International Conference on Robotics and Automation (ICRA)}, pages 811--817, 2019.

\bibitem[Gafur et~al.(2022)Gafur, Kanagalingam, Wagner, and Ruskowski]{Gafur2022-yi}
Nigora Gafur, Gajanan Kanagalingam, Achim Wagner, and Martin Ruskowski.
\newblock Dynamic collision and deadlock avoidance for multiple robotic manipulators.
\newblock \emph{IEEE Access}, 10:\penalty0 55766--55781, 2022.

\bibitem[Gammell et~al.(2020)Gammell, Barfoot, and Srinivasa]{Gammell2020-bd}
Jonathan~D Gammell, Timothy~D Barfoot, and Siddhartha~S Srinivasa.
\newblock Batch informed trees ({BIT*}): Informed asymptotically optimal anytime search.
\newblock \emph{International Journal of Robotics Research}, 39\penalty0 (5):\penalty0 543--567, April 2020.

\bibitem[Gao and Yu(2022)]{Gao2022-iu}
Kai Gao and Jingjin Yu.
\newblock Toward efficient task planning for dual-arm tabletop object rearrangement.
\newblock In \emph{Proceedings of the IEEE/RSJ International Conference on Intelligent Robots and Systems (IROS)}, pages 10425--10431, 2022.

\bibitem[Gao et~al.(2024{\natexlab{a}})Gao, Ye, Zhang, Huang, and Yu]{Gao2024-bl}
Kai Gao, Zihe Ye, Duo Zhang, Baichuan Huang, and Jingjin Yu.
\newblock Toward holistic planning and control optimization for dual-arm rearrangement.
\newblock \emph{aXiv 2404.06758}, 2024{\natexlab{a}}.

\bibitem[Gao et~al.(2024{\natexlab{b}})Gao, Zhaxizhuoma, Ding, Zhang, and Yu]{Gao2023-kx}
Kai Gao, Zhaxizhuoma, Yan Ding, Shiqi Zhang, and Jingjin Yu.
\newblock Orla*: Mobile manipulator-based object rearrangement with lazy a star.
\newblock \emph{arXiv 2309.13707}, 2024{\natexlab{b}}.

\bibitem[Gharbi et~al.(2009)Gharbi, Cortés, and Siméon]{Gharbi2009-rv}
Mokhtar Gharbi, Juan Cortés, and Thierry Siméon.
\newblock Roadmap composition for multi-arm systems path planning.
\newblock In \emph{Proceedings of the IEEE/RSJ International Conference on Intelligent Robots and Systems (IROS)}, pages 2471--2476, October 2009.

\bibitem[Gilday et~al.(2018)Gilday, Hughes, and Iida]{8593852}
Kieran Gilday, Josie Hughes, and Fumiya Iida.
\newblock Achieving flexible assembly using autonomous robotic systems.
\newblock In \emph{Proceedings of the IEEE/RSJ International Conference on Intelligent Robots and Systems (IROS)}, pages 1--9, 2018.

\bibitem[Harada et~al.(2014)Harada, Tsuji, and Laumond]{Harada2014-zp}
Kensuke Harada, Tokuo Tsuji, and Jean-Paul Laumond.
\newblock A manipulation motion planner for dual-arm industrial manipulators.
\newblock In \emph{Proceedings of the IEEE International Conference on Robotics and Automation (ICRA)}, pages 928--934, May 2014.

\bibitem[Hartmann et~al.(2023)Hartmann, Orthey, Driess, Oguz, and Toussaint]{Hartmann2021-kf}
Valentin~N. Hartmann, Andreas Orthey, Danny Driess, Ozgur~S. Oguz, and Marc Toussaint.
\newblock Long-horizon multi-robot rearrangement planning for construction assembly.
\newblock \emph{IEEE Transactions on Robotics}, 39\penalty0 (1):\penalty0 239--252, 2023.

\bibitem[Hoenig et~al.(2016)Hoenig, Kumar, Cohen, Ma, Xu, Ayanian, and Koenig]{Honig2016-ts}
Wolfgang Hoenig, T.~K. Kumar, Liron Cohen, Hang Ma, Hong Xu, Nora Ayanian, and Sven Koenig.
\newblock Multi-agent path finding with kinematic constraints.
\newblock In \emph{Proceedings of the International Conference on Automated Planning and Scheduling (ICAPS)}, volume~26, pages 477--485, Mar. 2016.

\bibitem[Honig et~al.(2019)Honig, Kiesel, Tinka, Durham, and Ayanian]{Honig2019-gr}
Wolfgang Honig, Scott Kiesel, Andrew Tinka, Joseph~W Durham, and Nora Ayanian.
\newblock Persistent and robust execution of {MAPF} schedules in warehouses.
\newblock \emph{IEEE Robotics and Automation Letters}, 4\penalty0 (2):\penalty0 1125--1131, April 2019.

\bibitem[Hönig et~al.(2018)Hönig, Preiss, Kumar, Sukhatme, and Ayanian]{Honig2018-zd}
Wolfgang Hönig, James~A Preiss, T~K~Satish Kumar, Gaurav~S Sukhatme, and Nora Ayanian.
\newblock Trajectory planning for quadrotor swarms.
\newblock \emph{IEEE Transactions on Robotics}, 34\penalty0 (4):\penalty0 856--869, August 2018.

\bibitem[Jiang et~al.(2023)Jiang, Wang, and Lu]{Jiang2023-cg}
Daqi Jiang, Hong Wang, and Yanzheng Lu.
\newblock Mastering the complex assembly task with a dual-arm robot: A novel reinforcement learning method.
\newblock \emph{IEEE Robotics and Automation Magazine}, 30\penalty0 (2):\penalty0 57--66, June 2023.

\bibitem[Jiang et~al.(2025)Jiang, Lin, and Li]{JiangAAAI25}
He~Jiang, Muhan Lin, and Jiaoyang Li.
\newblock Speedup techniques for switchable temporal plan graph optimization.
\newblock In \emph{Proceedings of the AAAI Conference on Artificial Intelligence (AAAI)}, 2025.

\bibitem[Koga and Latombe(1994)]{Koga1994-ty}
Y~Koga and J-C Latombe.
\newblock On multi-arm manipulation planning.
\newblock In \emph{Proceedings of the IEEE International Conference on Robotics and Automation (ICRA)}, pages 945--952 vol.2, 1994.

\bibitem[Kuffner and LaValle(2000)]{Kuffner2000-hn}
James~J. Kuffner and Steven~M. LaValle.
\newblock {RRT}-{Connect}: An efficient approach to single-query path planning.
\newblock In \emph{Proceedings of the IEEE International Conference on Robotics and Automation (ICRA)}, volume~2, pages 995--1001, 2000.

\bibitem[Li et~al.(2021)Li, Ruml, and Koenig]{li2021eecbs}
Jiaoyang Li, Wheeler Ruml, and Sven Koenig.
\newblock {EECBS}: A bounded-suboptimal search for multi-agent path finding.
\newblock In \emph{Proceedings of the AAAI Conference on Artificial Intelligence (AAAI)}, pages 12353--12362, 2021.

\bibitem[Liu et~al.(2023)Liu, Chen, Luo, and Liu]{liu2023simulation}
Ruixuan Liu, Alan Chen, Xusheng Luo, and Changliu Liu.
\newblock Simulation-aided learning from demonstration for robotic lego construction.
\newblock \emph{arXiv:2309.11010}, 2023.

\bibitem[Liu et~al.(2024{\natexlab{a}})Liu, Deng, Wang, and Liu]{Liu2024-go}
Ruixuan Liu, Kangle Deng, Ziwei Wang, and Changliu Liu.
\newblock Stablelego: Stability analysis of block stacking assembly.
\newblock \emph{IEEE Robotics and Automation Letters}, 9\penalty0 (11):\penalty0 9383--9390, 2024{\natexlab{a}}.

\bibitem[Liu et~al.(2024{\natexlab{b}})Liu, Sun, and Liu]{Liu2023-ny}
Ruixuan Liu, Yifan Sun, and Changliu Liu.
\newblock A lightweight and transferable design for robust lego manipulation.
\newblock In \emph{International Symposium on Flexible Automation}, page V001T07A004, 07 2024{\natexlab{b}}.

\bibitem[Liu et~al.(2025)Liu, Chen, Zhao, and Liu]{Liu2024-hk}
Ruixuan Liu, Alan Chen, Weiye Zhao, and Changliu Liu.
\newblock Physics-aware combinatorial assembly sequence planning using data-free action masking.
\newblock \emph{IEEE Robotics and Automation Letters}, 10\penalty0 (5):\penalty0 4882--4889, 2025.

\bibitem[Luo et~al.(2024)Luo, Xu, Liu, and Liu]{10552885}
Xusheng Luo, Shaojun Xu, Ruixuan Liu, and Changliu Liu.
\newblock Decomposition-based hierarchical task allocation and planning for multi-robots under hierarchical temporal logic specifications.
\newblock \emph{IEEE Robotics and Automation Letters}, 9\penalty0 (8):\penalty0 7182--7189, 2024.

\bibitem[Ma et~al.(2017{\natexlab{a}})Ma, Kumar, and Koenig]{Ma2017-wl}
Hang Ma, T.~K.~Satish Kumar, and Sven Koenig.
\newblock Multi-agent path finding with delay probabilities.
\newblock In \emph{Proceedings of the AAAI Conference on Artificial Intelligence (AAAI)}, page 3605–3612, 2017{\natexlab{a}}.

\bibitem[Ma et~al.(2017{\natexlab{b}})Ma, Li, Kumar, and Koenig]{Ma2017-iu}
Hang Ma, Jiaoyang Li, T.~K.~Satish Kumar, and Sven Koenig.
\newblock Lifelong multi-agent path finding for online pickup and delivery tasks.
\newblock In \emph{Proceedings of the 16th International Conference on Autonomous Agents and MultiAgent Systems (AAMAS)}, pages 837--845, 2017{\natexlab{b}}.

\bibitem[Maeda et~al.(2016)Maeda, Nakano, Maekawa, and Maruo]{7759340}
Yusuke Maeda, Ojiro Nakano, Takashi Maekawa, and Shoji Maruo.
\newblock From {CAD} models to toy brick sculptures: A {3D} block printer.
\newblock In \emph{Proceedings of the IEEE/RSJ International Conference on Intelligent Robots and Systems (IROS)}, pages 2167--2172, 2016.

\bibitem[Marcucci et~al.(2024)Marcucci, Umenberger, Parrilo, and Tedrake]{Marcucci2021-yv}
Tobia Marcucci, Jack Umenberger, Pablo~A. Parrilo, and Russ Tedrake.
\newblock Shortest paths in graphs of convex sets.
\newblock \emph{SIAM Journal on Optimization}, 34\penalty0 (1):\penalty0 507--532, 2024.

\bibitem[Meehan et~al.(2022)Meehan, Roberts, and Hiatt]{Meehan2022Asynchronous}
Charles~A Meehan, Mark Roberts, and Laura~M Hiatt.
\newblock Asynchronous motion planning and execution for a dual-arm robot.
\newblock In \emph{Workshop on Planning and Robotics held in conjunction with the International Conference on Automated Planning and Scheduling (ICAPS)}, Virtual, June 2022.

\bibitem[Michalos et~al.(2010)Michalos, Makris, Papakostas, Mourtzis, and Chryssolouris]{MICHALOS201081}
G.~Michalos, S.~Makris, N.~Papakostas, D.~Mourtzis, and G.~Chryssolouris.
\newblock Automotive assembly technologies review: challenges and outlook for a flexible and adaptive approach.
\newblock \emph{{CIRP} Journal of Manufacturing Science and Technology}, 2\penalty0 (2):\penalty0 81--91, 2010.

\bibitem[Mirrazavi~Salehian et~al.(2018)Mirrazavi~Salehian, Figueroa, and Billard]{Mirrazavi-Salehian2018-jm}
Seyed~Sina Mirrazavi~Salehian, Nadia Figueroa, and Aude Billard.
\newblock A unified framework for coordinated multi-arm motion planning.
\newblock \emph{International Journal of Robotics Research}, 37\penalty0 (10):\penalty0 1205--1232, September 2018.

\bibitem[Nägele et~al.(2020)Nägele, Hoffmann, Schierl, and Reif]{9341428}
Ludwig Nägele, Alwin Hoffmann, Andreas Schierl, and Wolfgang Reif.
\newblock Legobot: Automated planning for coordinated multi-robot assembly of lego structures.
\newblock In \emph{Proceedings of the IEEE/RSJ International Conference on Intelligent Robots and Systems (IROS)}, pages 9088--9095, 2020.

\bibitem[of~Standards and Technology(2024)]{nist-box}
National~Institute of~Standards and Technology.
\newblock {STEP} {File} {Viewer} - {Box} assembly, 2024.
\newblock URL \url{https://pages.nist.gov/step-file-viewer.html}.

\bibitem[Okumura and D\'{e}fago(2023)]{Okumura2022-mw}
Keisuke Okumura and Xavier D\'{e}fago.
\newblock Quick multi-robot motion planning by combining sampling and search.
\newblock In \emph{Proceedings of the Thirty-Second International Joint Conference on Artificial Intelligence (IJCAI)}, 2023.

\bibitem[Pan et~al.(2021)Pan, Wells, Shome, and Kavraki]{Pan2021-ar}
Tianyang Pan, Andrew~M Wells, Rahul Shome, and Lydia~E Kavraki.
\newblock A general task and motion planning framework for multiple manipulators.
\newblock In \emph{Proceedings of the IEEE/RSJ International Conference on Intelligent Robots and Systems (IROS)}, pages 3168--3174, September 2021.

\bibitem[Popov et~al.(2017)Popov, Heess, Lillicrap, Hafner, Barth-Maron, Vecerik, Lampe, Tassa, Erez, and Riedmiller]{popov2017dataefficient}
Ivaylo Popov, Nicolas Heess, Timothy Lillicrap, Roland Hafner, Gabriel Barth-Maron, Matej Vecerik, Thomas Lampe, Yuval Tassa, Tom Erez, and Martin Riedmiller.
\newblock Data-efficient deep reinforcement learning for dexterous manipulation.
\newblock \emph{arXiv 1704.03073}, 2017.

\bibitem[Shaoul et~al.(2024{\natexlab{a}})Shaoul, Mishani, Likhachev, and Li]{shaoulmishani2024xcbs}
Yorai Shaoul, Itamar Mishani, Maxim Likhachev, and Jiaoyang Li.
\newblock Accelerating search-based planning for multi-robot manipulation by leveraging online-generated experiences.
\newblock In \emph{Proceedings of International Conference on Automated Planning and Scheduling (ICAPS)}, 2024{\natexlab{a}}.

\bibitem[Shaoul et~al.(2024{\natexlab{b}})Shaoul, Veerapaneni, Likhachev, and Li]{shaoul2024gencbs}
Yorai Shaoul, Rishi Veerapaneni, Maxim Likhachev, and Jiaoyang Li.
\newblock Unconstraining multi-robot manipulation: Enabling arbitrary constraints in ecbs with bounded sub-optimality.
\newblock In \emph{Proceedings of International Symposium on Combinatorial Search (SoCS)}, 2024{\natexlab{b}}.

\bibitem[Shome and Bekris(2019)]{Shome2019-ak}
Rahul Shome and Kostas~E. Bekris.
\newblock Anytime multi-arm task and motion planning for pick-and-place of individual objects via handoffs.
\newblock In \emph{Proceedings of International Symposium on Multi-Robot and Multi-Agent Systems (MRS)}, pages 37--43, 2019.

\bibitem[Shome and Bekris(2021)]{Shome2021-ka}
Rahul Shome and Kostas~E. Bekris.
\newblock Synchronized multi-arm rearrangement guided by mode graphs with capacity constraints.
\newblock In \emph{Algorithmic Foundations of Robotics XIV}, pages 243--260, Cham, 2021. Springer International Publishing.

\bibitem[Shome et~al.(2020)Shome, Solovey, Dobson, Halperin, and Bekris]{Shome2020-hc}
Rahul Shome, Kiril Solovey, Andrew Dobson, Dan Halperin, and Kostas~E Bekris.
\newblock {dRRT*}: Scalable and informed asymptotically-optimal multi-robot motion planning.
\newblock \emph{Autononomus Robots}, 44\penalty0 (3):\penalty0 443--467, March 2020.

\bibitem[Shome et~al.(2021)Shome, Solovey, Yu, Bekris, and Halperin]{Shome2021-ke}
Rahul Shome, Kiril Solovey, Jingjin Yu, Kostas Bekris, and Dan Halperin.
\newblock Fast, high-quality two-arm rearrangement in synchronous, monotone tabletop setups.
\newblock \emph{IEEE Transactions on Automation Science and Engineering}, 18\penalty0 (3):\penalty0 888--901, 2021.

\bibitem[Smith et~al.(2012{\natexlab{a}})Smith, Karayiannidis, Nalpantidis, Gratal, Qi, Dimarogonas, and Kragic]{SMITH20121340}
Christian Smith, Yiannis Karayiannidis, Lazaros Nalpantidis, Xavi Gratal, Peng Qi, Dimos~V. Dimarogonas, and Danica Kragic.
\newblock Dual arm manipulation -- {A} survey.
\newblock \emph{Robotics and Autonomous Systems}, 60\penalty0 (10):\penalty0 1340--1353, 2012{\natexlab{a}}.

\bibitem[Smith et~al.(2012{\natexlab{b}})Smith, Karayiannidis, Nalpantidis, Gratal, Qi, Dimarogonas, and Kragic]{Smith2012-if}
Christian Smith, Yiannis Karayiannidis, Lazaros Nalpantidis, Xavi Gratal, Peng Qi, Dimos~V Dimarogonas, and Danica Kragic.
\newblock Dual arm manipulation—a survey.
\newblock \emph{Robotics and Autonomous Systems}, 60\penalty0 (10):\penalty0 1340--1353, October 2012{\natexlab{b}}.

\bibitem[Solis et~al.(2021)Solis, Motes, Sandström, and Amato]{Solis2021-aj}
Irving Solis, James Motes, Read Sandström, and Nancy~M Amato.
\newblock Representation-optimal multi-robot motion planning using conflict-based search.
\newblock \emph{IEEE Robotics and Automation Letters}, 6\penalty0 (3):\penalty0 4608--4615, July 2021.

\bibitem[Solovey et~al.(2016)Solovey, Salzman, and Halperin]{Solovey2016-dz}
Kiril Solovey, Oren Salzman, and Dan Halperin.
\newblock Finding a needle in an exponential haystack: Discrete {RRT} for exploration of implicit roadmaps in multi-robot motion planning.
\newblock \emph{International Journal of Robotics Research}, 35\penalty0 (5):\penalty0 501--513, April 2016.

\bibitem[Stern et~al.(2019)Stern, Sturtevant, Felner, Koenig, Ma, Walker, Li, Atzmon, Cohen, Kumar, Boyarski, and Bart{\'{a}}k]{Stern2019-ck}
Roni Stern, Nathan~R. Sturtevant, Ariel Felner, Sven Koenig, Hang Ma, Thayne~T. Walker, Jiaoyang Li, Dor Atzmon, Liron Cohen, T.~K.~Satish Kumar, Eli Boyarski, and Roman Bart{\'{a}}k.
\newblock Multi-agent pathfinding: Definitions, variants, and benchmarks.
\newblock In \emph{Proceedings of the Twelfth Annual Symposium on Combinatorial Search (SoCS)}, pages 151--159, 2019.

\bibitem[Stoop et~al.(2023)Stoop, Ratnayake, and Toffetti]{Stoop2023-im}
Pascal Stoop, Tharaka Ratnayake, and Giovanni Toffetti.
\newblock A method for multi-robot asynchronous trajectory execution in \textit{MoveIt2}.
\newblock \emph{arXiv 2310.08597}, 2023.

\bibitem[Su et~al.(2024)Su, Veerapaneni, and Li]{Su2023-jp}
Yifan Su, Rishi Veerapaneni, and Jiaoyang Li.
\newblock Bidirectional temporal plan graph: enabling switchable passing orders for more efficient multi-agent path finding plan execution.
\newblock In \emph{Proceedings of the AAAI Conference on Artificial Intelligence (AAAI)}, 2024.

\bibitem[Sucan et~al.(2012)Sucan, Moll, and Kavraki]{OMPL2012}
Ioan~A. Sucan, Mark Moll, and Lydia~E. Kavraki.
\newblock The open motion planning library.
\newblock \emph{IEEE Robotics and Automation Magazine}, 19\penalty0 (4):\penalty0 72--82, 2012.

\bibitem[Tian et~al.(2022)Tian, Xu, Li, Luo, Sueda, Li, Willis, and Matusik]{10.1145/3550454.3555525}
Yunsheng Tian, Jie Xu, Yichen Li, Jieliang Luo, Shinjiro Sueda, Hui Li, Karl D.~D. Willis, and Wojciech Matusik.
\newblock Assemble them all: Physics-based planning for generalizable assembly by disassembly.
\newblock \emph{ACM Transactions on Graphics}, 41\penalty0 (6):\penalty0 278:1--278:11, 2022.

\bibitem[Varambally et~al.(2022)Varambally, Li, and Koenig]{Varambally2022-ux}
Sumanth Varambally, Jiaoyang Li, and Sven Koenig.
\newblock Which {MAPF} model works best for automated warehousing?
\newblock In \emph{Proceedings of International Symposium on Combinatorial Search (SoCS)}, volume~15, pages 190--198, July 2022.

\bibitem[Xian et~al.(2017)Xian, Lertkultanon, and Pham]{Xian2017-fz}
Zhou Xian, Puttichai Lertkultanon, and Quang-Cuong Pham.
\newblock Closed-chain manipulation of large objects by multi-arm robotic systems.
\newblock \emph{IEEE Robotics and Automation Letters}, 2\penalty0 (4):\penalty0 1832--1839, October 2017.

\bibitem[Zhang and Pecora(2024)]{Zhang_undated-hg}
Shiyu Zhang and Federico Pecora.
\newblock Online and scalable motion coordination for multiple robot manipulators in shared workspaces.
\newblock \emph{IEEE Transactions on Automation Science and Engineering}, PP\penalty0 (99):\penalty0 1--20, 2024.

\end{thebibliography}

\newpage
\clearpage
\appendix
\subsection{Details of Assembly Sequence Planning}
\label{sec:app-asp}

Given a \LEGO{} design with $N_a$ objects, we auto-generate the assembly plan $A=[a_1, a_2, \dots, a_{N_a}]$.
In particular, we use assemble-by-disassemble (AbD) \cite{10.1145/3550454.3555525} and start from the fully assembled structure $\{a_1, a_2, \dots, a_{N_a}\}$.
For each step, we pick an action $a_j$ and remove it from the structure.
We keep removing bricks until we have nothing remained.
The assembly plan $A$ is then generated by reversing the order of removing bricks from the structure.
Specifically, we use a depth-first search tree (DFS) to implement the AbD search.
To ensure the assembly sequence is physically executable, we need to ensure (1) the intermediate structure is stable, (2) the structure is stable under robot operation, and (3) robots have sufficient operating space.
Thus, we use the action mask in \cite{Liu2024-hk} to validate the selection of $a_j$ at each step and prune the tree branches.
The action mask evaluates (3) by discretizing the environment and checking the occupancy grids.
It estimates (1) and (2) by leveraging the stability analysis in \cite{Liu2024-go} since it can efficiently estimate the static physical stability of a given \LEGO{} structure as well as dynamic stability under external weights.
Note that generating $A$ using AbD is more efficient compared to forward search since it results in a significantly smaller search space after applying the action mask for branch pruning. 
For general assembly tasks, physics engine-guided AbD search \cite{10.1145/3550454.3555525} can similarly generate assembly plans.

\subsection{Details of Task Planning Formulation}
\label{sec:app-task}

Each assembly step $a_j$ requires a specific LEGO type, and each brick in the environment has a specific type. Typically, there are many bricks of the same type for in the environment. 
Mathematically, we represent whether a brick $b_k$ is the correct type for the assembly step $a_j$ using a binary variable $\delta^t_{j,k}$, where $\delta^t_{j,k} = 1$ if $b_k$ is suitable and $\delta^t_{j,k} = 0$ otherwise.
The assembly sequence also specifies whether each step $a_j$ requires multiple robots for cooperative assembly, denoted by another binary variable $\delta^s_{j}$, where $\delta^s_{j} = 1$ if two robots are needed for $a_j$. There are two cases when a support arm may be required. If the brick is assembled from the bottom, a handover is always required to reorient the block and a support from top is necessary during assembly.

Let $X_{ijkg}$ and $Y_{ijg}$ be the binary decision variables. $X_{ijkg = 1}$ means that robot $i$ is assigned to assembly step $a_j$, using object $b_k$ and the $g^{th}$ grasp pose. $Y_{ijg} = 1$ means that the $i^{th}$ robot is assigned to support the assembly step $j$, using the $g^{th}$ support pose. $C_{ijkg}$ and $C^s_{ijg}$ denote the estimated cost for assigning $X_{ijkg}=1$ and $Y_{ijg}=1$, respectively.
$P$ is the maximum number of feasible grasp poses and is set based on the objects and user preferences. 
$X_{ijkg}=1$ means robot $i$ and object $b_k$ are assigned for the assembly step $j$, using the $g^{th}$ grasp pose. 
Cost $C_{ijkg}$ is computed as the sum of joint-space distance from robot $i$'s \home{} pose to $g^{th}$ grasp pose for picking object $b_k$ and then to the robot target pose for placing the object at assembly step $a_j$. 
Similarly, $C^s_{ijg}$ is the distance from robot $i$'s \home{} pose, to the $g^{th}$ support pose for step $a_j$. 
For tasks that require handover, costs are computed differently since the primary robot will receive the LEGO brick and assemble the robot, while the support robot is responsible for first picking up the robot, handover to the primary robot, and supporting the structure from the top.  
Specifically, the cost of the primary robot $C_{ijkg}$ is the sum of the cost for the support robot\footnote{We assume there is a single, known robot to pick and handover. For dual-arm LEGO assembly, the support robot is always the other robot.} to pick $b_k$ with the grasp pose $g^{th}$ and the cost of robot $i$ to receive and place $b_k$. 
The cost of the support robot $C^s_{ijg}$ is the sum of the cost to handover the brick and support the structure.
The solution to the following ILP program then forms a complete sequential task plan.

\begin{align}
\argmin_{X, Y} &\sum_{ijkg} C_{ijkg} X_{ijkg} + \sum_{ijg} C^s_{ijg} Y_{ijg} + \lambda \sum_{j} (Z^M_j - Z^m_j) \label{eq:assign}\\
\text{subect to} & \text{~task, robot, and object constraints} \nonumber
\end{align} 

The first two terms of optimization minimize the total cost of executing a sequential task plan. The last auxiliary cost term penalizes cases where a single robot is assigned for multiple consecutive tasks and prevents parallelization. The constraints include
\begin{align}
  \sum_{i=1}^{N} \sum_{k=1}^{N_b} \sum_{g=1}^{P} X_{ijkg} &= 1 \quad \forall j \in [1, \dots, N_a]\label{eq:task_once} \\
  \sum_{i=1}^{N} \sum_{g=1}^{P} Y_{ijg} &= \delta^s_{j} \quad \forall j  \label{eq:sup_once}\\
  \sum_{k=1}^{N_b} \sum_{g=1}^{P} X_{ijkg} + \sum_{g=1}^{P} Y_{ijg} &\leq 1 \quad \forall i \in [1, \dots, N], \; \forall j \label{eq:robot_once}\\
  \sum_{i=1}^{N} \sum_{k=1}^{N_b} \sum_{g=1}^{P} X_{ijkg} \delta^t_{jk} &= 1 \quad \forall j  \label{eq:obj_match}\\
 \sum_{i=1}^{N} \sum_{j=1}^{N_a} \sum_{g=1}^{P} X_{ijkg} &\leq 1 \quad \forall k \in [1, \dots N_b] \label{eq:object_once}
\end{align}

Eqn. \ref{eq:task_once} ensures that each task is assigned exactly one primary robot, one object, and one grasp pose. Eqn. \ref{eq:sup_once} assigns a second support robot and a support pose to each assembly step that needs one. Eqn. \ref{eq:robot_once} prevents each robot from being the primary robot and the support robot in the same step. Eqn. \ref{eq:obj_match} matches an object of the correct type for each assembly step. Eqn. \ref{eq:object_once} ensures that each object is used at most once. Furthermore, the following constraints apply to the auxiliary variables for all $ i \in [1, \dots, N]$ and $ j \in [1, \dots, N_a-N+1] $

\begin{align}
z_{ij} &= \sum_{j'=j}^{j+N-1} \sum_{k=1}^{N_b} \sum_{g=1}^{P} X_{ij'tg} + \sum_{j'=j}^{j+N-1} \sum_{g=1}^{P} Y_{ij'g} \label{eq:zaux} \\
Z^M_j &\geq z_{ij}, \quad Z^m_j \leq z_{ij} \label{eq:zmax}
\end{align}

$z_{ij}$ denotes the number of tasks assigned to robot $i$ from the window of assembly steps $a_j$ to $a_{j+N-1}$, whereas $Z^M_j$ and $Z^m_j$ are the maximum and minimum $z_{ij}$ at each window for all robot $i$.  The window size is set to the same as the number of robots. 

\subsection{Details of LEGO Manipulation Policy}\label{sec:app-skill}
Our \LEGO{} manipulation policy is composed of 6 manipulation skills as shown in \cref{fig:dual_arm_example}.
For the learned force policy (\ie pick, place-down, handover, place-up), the robot follows a two-step motion sequence, \ie \textit{attach} and \textit{twist}, to manipulate the \LEGO{} brick. 
The twist angle and axis of the EOAT are learned with a safe-learning framework following \cite{Liu2023-ny} to ensure the EOAT can successfully attach or release the \LEGO{} brick. 
For goal-reaching force policy (\ie support-up, support-down), the robot moves to its target pose in a one-step motion.

\apex~requires a reference path for each manipulation policy to check collisions in the TPG construction and shortcutting process.  
Thus, for each learned force policy, we generate a sequence of subgoals (\ie robot target pose) that corresponds to each of the two steps, \textit{attach} and \textit{twist}.
Algorithmically, the EOAT pose in \textit{attach} subgoal is calculated based on the known \LEGO{} pose in the assembly plate, the specific press side, and the press pose offset for the \LEGO{} brick. 
The \textit{twist} is set by rotating the \textit{attach} subgoal with the learned twist parameters.
Then, the two joint space subgoals are computed with inverse kinematics for each robot to determine their feasibility. 
The target pose of the previous transit task is computed by adding a small z-offset (\eg 5mm) to the \textit{attach} subgoal, where the z-axis is orthogonal to the assembly plate. The reference path for planning is computed with RRT-connect. 

For support-up, the EOAT is set to support from below the \LEGO{} brick in the current substructure that would sit below the current brick that is being placed down. Similarly for support-down pose, the EOAT is set to support from the top of the \LEGO{} brick in the current substructure that sits above the current brick that is being placed-up. Then, the feasibility of these goal-reaching policies can be determined from the inverse kinematic for each robot. Also, the goal pose of the previous transit task is computed by adding a small z-offset to the support pose. 

Since the contact models of \LEGO{} bricks are not directly simulated, these manipulation skills are treated as a (sequence of) motion primitive(s) and directly executed, similar to transit tasks in simulation. 
In the real bimanual manipulation setup, force feedback determines contact location and successful attach positions. 
Specifically, for the learned force policy (\ie pick, place-down, handover, place-up), the robot EOAT moves toward the direction of the \textit{attach} pose in the z-axis until a predetermined force threshold (\eg 15N). 
Once this force threshold is reached, the stopping point determines the new \textit{twist} axis, and a new \textit{twist} pose is calculated for the robot. 
For the goal-reaching policy (\ie support-up, support-down), the robot EOAT moves in the direction of the \textit{support} pose until a small predetermined force threshold is reached (\eg 0.1N), and the stopping point becomes the actual support pose. 
The robot moves at low speed during the support skills to ensure that our robot can stop in time ($\leq 30ms$) to avoid damaging the stability of the existing structure. 

\begin{figure}[t!]
\centering
\subfigure[]{\includegraphics[width=0.31\linewidth]{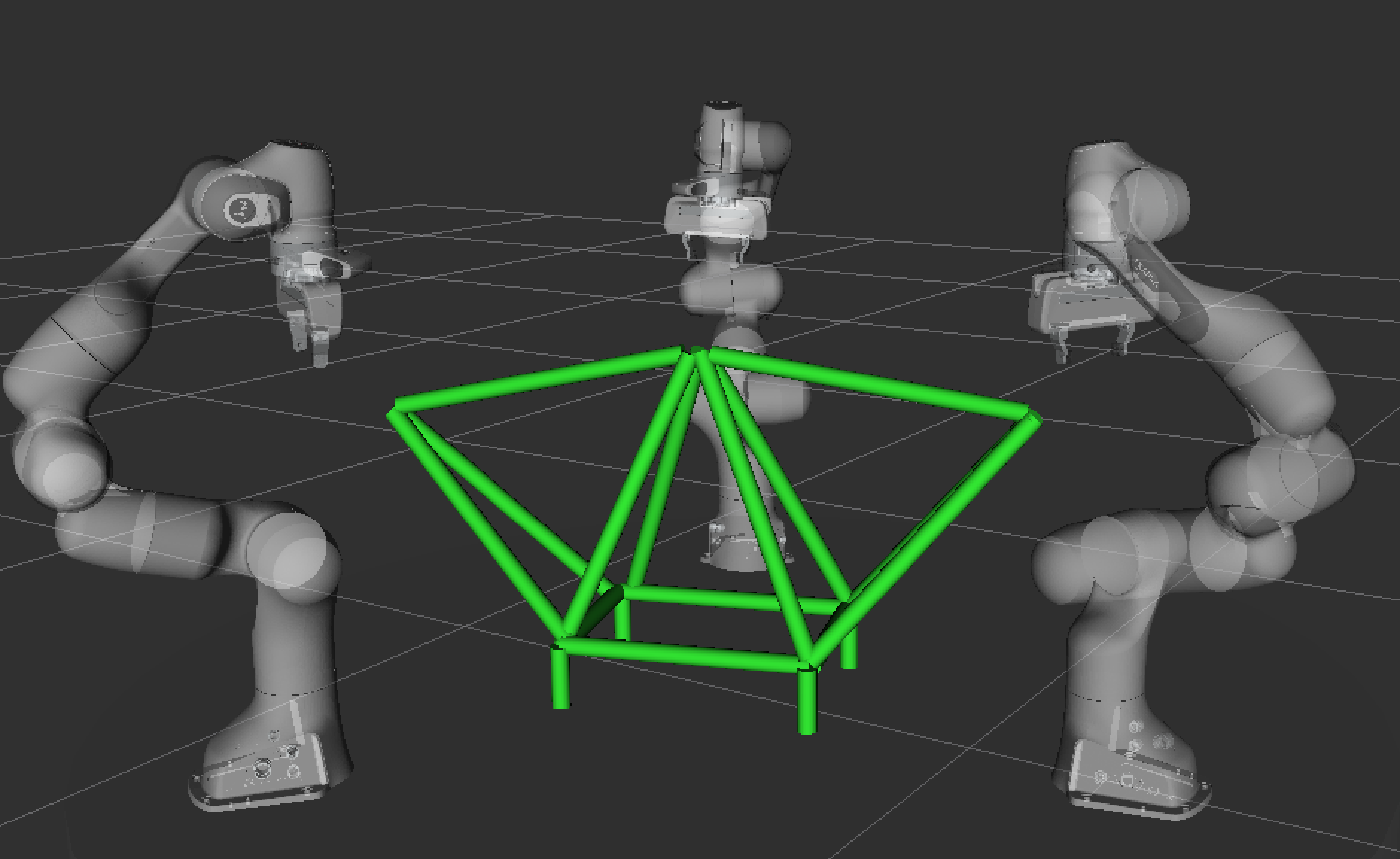}}\label{fig:panda_three_boat}\hfill
\subfigure[]{\includegraphics[width=0.31\linewidth]{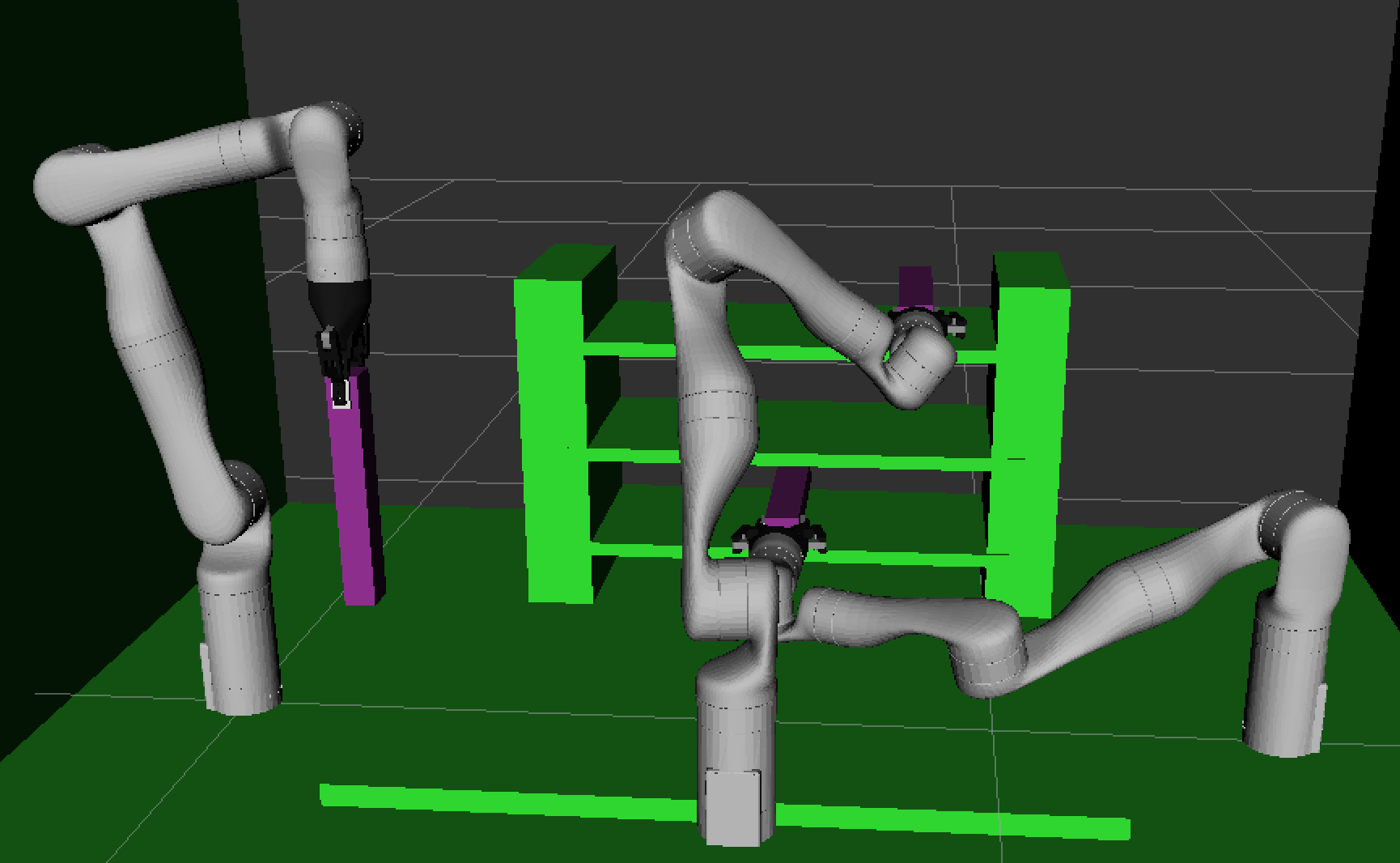}}\label{fig:kinova_three}\hfill
\subfigure[]{\includegraphics[width=0.31\linewidth]{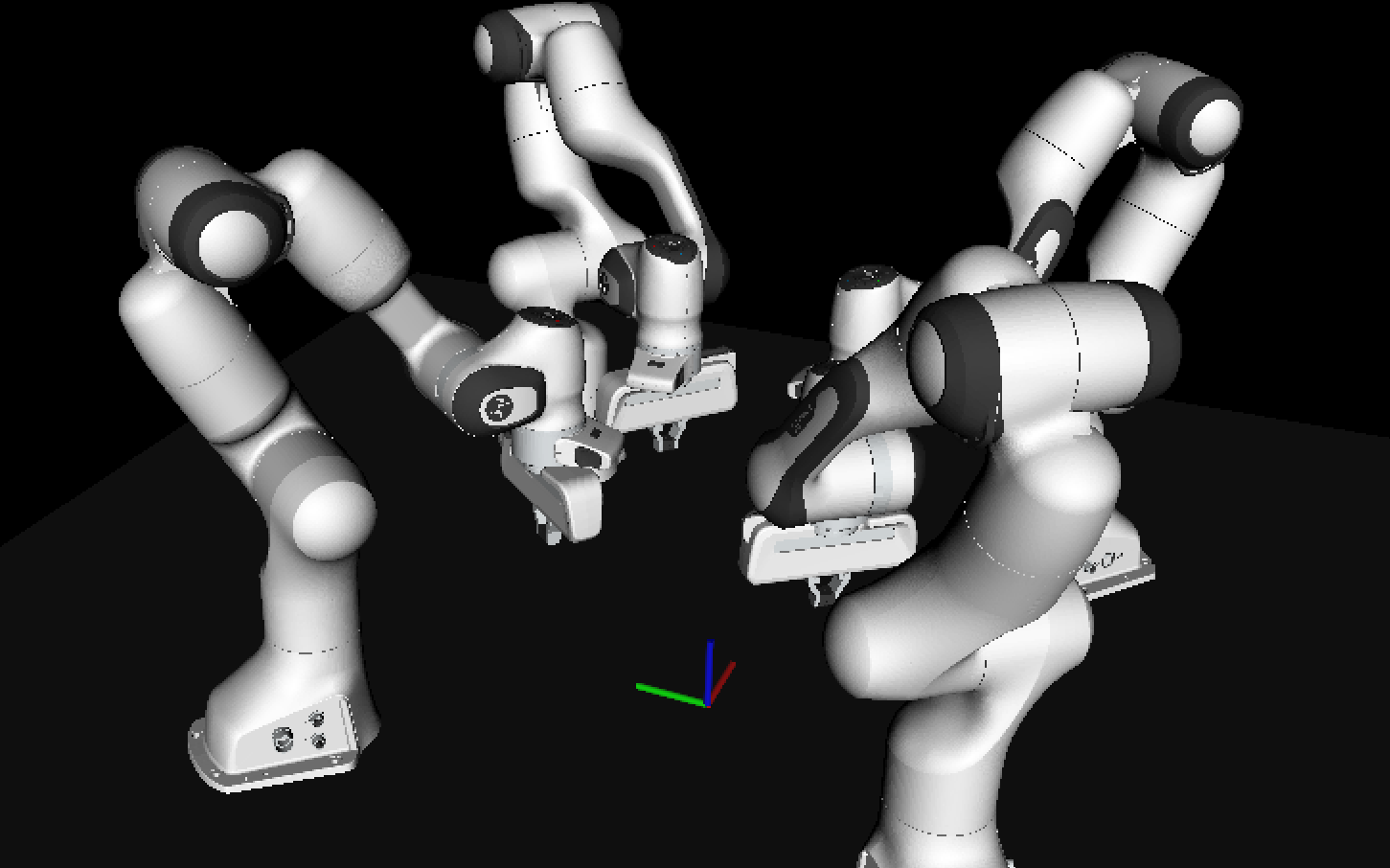}}\label{fig:panda_four}
    \caption{Experiment setup for additional TPG construction time evaluation. (a) Three Panda arm assembly. (b) Three Kinova arm rearrangement. (c) Four Panda arm multi-goal motion.  \label{fig:additional_scale}}
\end{figure}

\begin{table}[t!]
\centering
\caption{TPG construction time for additional multi-robot experiments in Fig. \ref{fig:additional_scale}.}
\label{tab:tpg_runtime_summary}
\begin{tabular}{lccc}
\toprule
Scenario & \# Robots & \makecell{Motion Plan \\ Duration (s)} & \makecell{TPG Construction\\ Time (s)} \\
\midrule
(a) Assembly & 3 & 98 & 23 \\
(b) Rearrangement & 3 & 70 & 70 \\
(c) Multi-Goal Motion & 4 & 52 & 78 \\
\bottomrule
\end{tabular}
\end{table}

\begin{figure*}
\includegraphics[width=\textwidth]{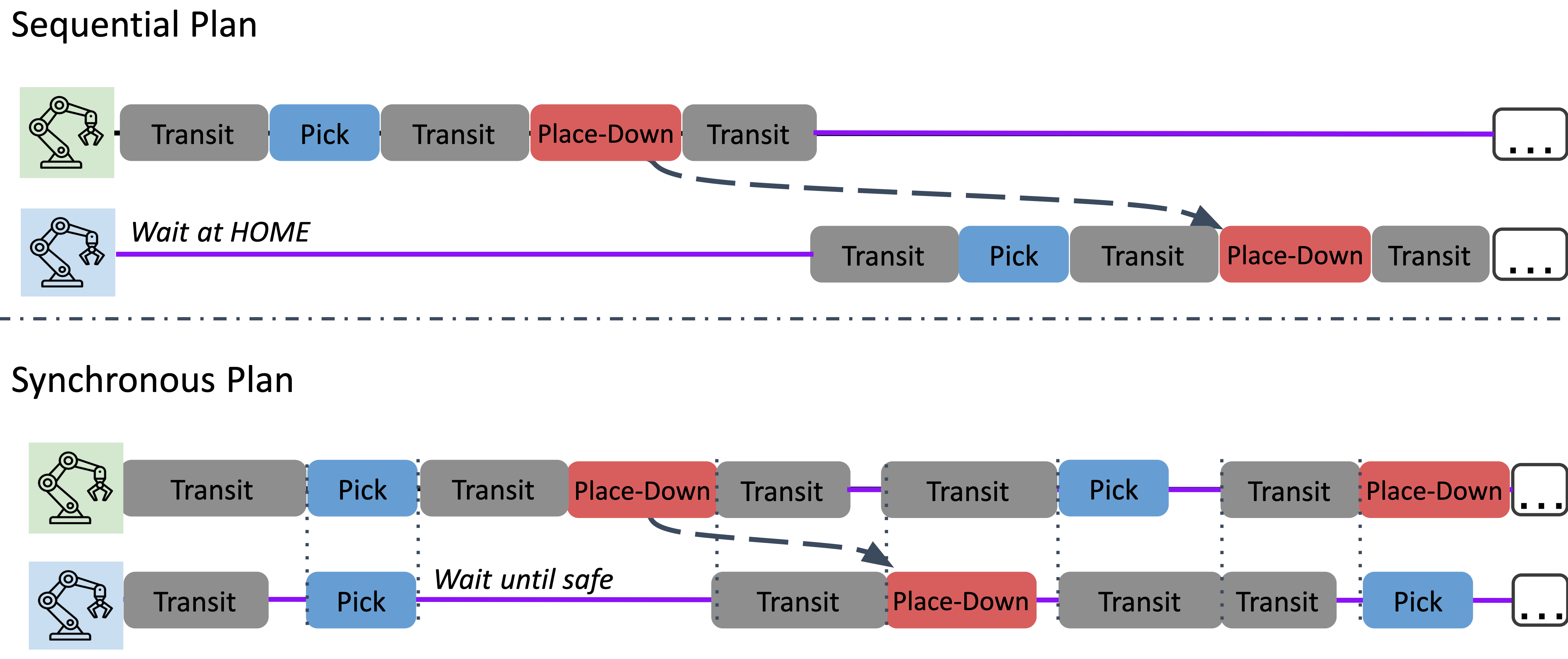} 
\caption{Converting a sequential task plan to a synchronous task plan. The sequential task plan produces a turn-based, sequential motion plan, whereas in the synchronous task plan, each robot waits for other robots to complete the current task before starting the next task at the same time. If it is unsafe to complete two tasks simultaneously due to collision, tasks scheduled later in the sequential plan would wait until the earlier robot completes its task, and it becomes safe to proceed simultaneously. }
\label{fig:seq_v_sync}
\end{figure*}

\begin{table*}[h!]
        \centering
\caption{Wall clock time for planning, TPG construction, and shortcutting across environments (mean ± std)}
\label{tab:planning_metrics}
\setlength{\tabcolsep}{4pt} 
\begin{tabular}{@{}ccccccccccc@{}}
\toprule
& \textbf{Version} & \textbf{Cliff}& \textbf{Branched Stairs}& \textbf{Faucet}& \textbf{Bridge}& \textbf{Fish}& \textbf{Chair}& \textbf{Vessel}& \textbf{Guitar}& \textbf{RSS}\\ 
\# Objects & & 11& 14& 14& 38& 29& 258& 36& 24& 47\\
\midrule
\multirow{2}{*}{\makecell[c]{Task \\ Planning}} & \apex{} & \textbf{0.6 ± 0.0} & \textbf{1.1 ± 0.0} & \textbf{1.1 ± 0.0} & \textbf{9.7 ± 0.5} & \textbf{6.5 ± 0.2} & \textbf{13.0 ± 0.2}  & \textbf{4.9 ± 0.6} & \textbf{2.3 ± 0.1} & \textbf{14.5 ± 0.5} \\
& Sync & 1.5 ± 0.0 & 1.9 ± 0.0 & 2.3 ± 0.0 & 10.8 ± 0.5 & 8.8 ± 0.2 & 27.1 ± 0.3 & 6.7 ± 0.5 & 3.9 ± 0.0 & 18.1 ± 0.5\\
\midrule
\multirow{2}{*}{\makecell[c]{Motion \\ Planning}} & \apex{} & \textbf{11.9 ± 0.3} & \textbf{14.7 ± 0.3} & \textbf{14.3 ± 0.2} & \textbf{52.9 ± 0.8} & \textbf{37.8 ± 0.6} & \textbf{220.3 ± 3.5} & \textbf{30.5 ± 0.7} & \textbf{20.2 ± 0.1} & \textbf{72.5 ± 4.6} \\
& Sync & 26.7 ± 0.5 & 36.4 ± 1.2 & 37.2 ± 0.5 & 124.8 ± 2.8 & 96.7 ± 1.0 & 501.9 ± 5.7 & 77.4 ± 0.8 & 46.8 ± 0.3 & 180.9 ± 7.6 \\
\midrule
\multirow{2}{*}{\makecell[c]{TPG \\ Construction}} & \apex{} & \textbf{9.1 ± 0.0} & \textbf{9.6 ± 0.1} & \textbf{10.3 ± 0.1} & 35.7 ± 0.6 & 29.2 ± 0.1 & \textbf{341.9 ± 1.4} & \textbf{16.2 ± 0.8} & \textbf{16.7 ± 0.1} & \textbf{52.2 ± 0.4} \\
& Sync & 9.4 ± 0.2 & 9.7 ± 0.0 & 10.7 ± 0.3 & \textbf{30.0 ± 0.6} & \textbf{27.0 ± 0.2} & 345.5 ± 4.3 & 16.8 ± 0.2 & 16.8 ± 0.0 & 53.8 ± 1.0 \\
\midrule
\multirow{2}{*}{\makecell[c]{Total \\ Planning}} & \apex{} & \textbf{21.6 ± 0.4} & \textbf{25.3 ± 0.4} & \textbf{25.7 ± 0.2} & \textbf{98.2 ± 1.8} & \textbf{73.5 ± 0.9} & \textbf{575.2 ± 5.1} & \textbf{51.6 ± 2.1} & \textbf{9.2 ± 0.2} & \textbf{139.3 ± 5.5} \\
& Sync & 37.5 ± 0.8 & 48.0 ± 1.2 & 50.1 ± 0.8 & 165.6 ± 3.8 & 132.5 ± 1.4 & 874.6 ± 10.3 & 100.9 ± 1.5 & 67.5 ± 0.3 & 252.8 ± 9.1\\
\midrule
TPG Shortcut & & 20& 20& 20& 60& 60& 60& 60& 20& 60\\
\bottomrule
        \end{tabular}
\end{table*}

\subsection{Additional Evaluation of TPG Construction Time}
\label{sec:app-exp}

To evaluate the scalability of TPG construction with repsecto to the number of robots, we conduct additional experiments in simulation involving scenarios with more than two robot arms. We examine three different cases:
\begin{itemize}
    \item Assembly task (Fig. \ref{fig:additional_scale} (a)): This scenario involves a 17-step assembly task where three Panda robot arms construct a boat-shaped structure using long rods, based on the task setup in~\citep{chen2022cooperativeMRAMP}. A sequential, 98-second motion plan is pre-computed using RRT-Connect. For this scenario, the TPG construction takes 23 seconds.
    \item Rearrangement task (Fig. \ref{fig:additional_scale} (b)): We test a rearrangement task with three Kinova robot arms simultaneously repositioning three long rods onto a shelf. Tasks are pre-assigned to each arm, and a 70-second synchronous multi-robot motion plan is pre-computed using xECBS \citep{shaoulmishani2024xcbs}. TPG construction finishes in 70 seconds.
    \item Multi-Goal Motion (Fig. \ref{fig:additional_scale} (c)): This scenario features a multi-goal motion planning problem with four Panda robot arms. The robots navigate sequentially through five intermediate goals according to a 52-second synchronous motion plan generated via RRT. The corresponding TPG construction time is 78 seconds.
\end{itemize}

These results are summarized in Table \ref{tab:tpg_runtime_summary}. Although collision checking complexity scales quadratically with the number of robots, the TPG construction time remains reasonable for systems involving 3-4 arms. It is worth noting that these experiments utilized exact robot geometries for collision checking to accommodate close-proximity cooperation. The construction time can be significantly reduced by using simplified geometric models.

\subsection{Illustration of a Synchronous Plan}
\label{sec:app-seq_v_sync}

 Fig. \ref{fig:seq_v_sync} shows a graphical illustration of how a synchronous task plan compares to a sequential task plan.

\subsection{Numerical Results of Simulation Experiments }
\label{sec:app-metrics}

 Table \ref{tab:planning_metrics} shows the wall clock time for planning, TPG construction, and shortcutting across environments averaged over 4 random seeds.

\end{document}